\documentclass{article}

\usepackage{microtype}
\usepackage{graphicx}

\usepackage{booktabs} 

\usepackage{caption}
\usepackage{subcaption}

\usepackage{hyperref}

\usepackage[preprint]{icml2026}

\usepackage{amsmath}
\usepackage{amssymb}
\usepackage{mathtools}
\usepackage{amsthm}
\usepackage{booktabs}
\usepackage{multicol}
\usepackage{booktabs}
\usepackage{multirow}
\usepackage{booktabs}
\usepackage[table]{xcolor}
\usepackage{longtable}

\usepackage[capitalize,noabbrev]{cleveref}

\theoremstyle{plain}

\theoremstyle{definition}

\theoremstyle{remark}

\usepackage[textsize=tiny]{todonotes}

\usepackage{listings}
\usepackage{xcolor}
\usepackage{tikz}
\usetikzlibrary{shapes.geometric, positioning, calc, fit, arrows.meta, shadows}

\begin{document}

\makeatletter
\renewcommand{\@pa}[1]{%
  \ifcsname the@affil#1\endcsname
  \else
    \stepcounter{@affiliationcounter}%
    \newcounter{@affil#1}%
    \setcounter{@affil#1}{\value{@affiliationcounter}}%
  \fi%
  \ifcsname @icmlsymbol#1\endcsname
    \textsuperscript{\csname @icmlsymbol#1\endcsname\,}%
  \else
    \textsuperscript{\arabic{@affil#1}\,}%
  \fi
}
\makeatother

\makeatletter
\renewcommand{\printAffiliationsAndNotice}[1]{\global\icml@noticeprintedtrue%
  \stepcounter{@affiliationcounter}%
  {\let\thefootnote\relax\footnotetext{\hspace*{-1.5\footnotesep}\ificmlshowauthors #1\fi%
      \forloop{@affilnum}{1}{\value{@affilnum} < \value{@affiliationcounter}}{
        \textsuperscript{†}%
        \ifcsname @affilname\the@affilnum\endcsname%
          \csname @affilname\the@affilnum\endcsname%
        \else
          {\bf AUTHORERR: Missing \textbackslash{}icmlaffiliation.}
        \fi
      }.%
      \ifdefined\icmlcorrespondingauthor@text
         { }Correspondence to: \icmlcorrespondingauthor@text.
      \else
        {\bf AUTHORERR: Missing \textbackslash{}icmlcorrespondingauthor.}
      \fi

      \ \\
      \Notice@String
    }
  }
}
\makeatother

\twocolumn[

\icmltitle{Scaling Behavior Cloning Improves Causal Reasoning: An Open Model for Real-Time Video Game Playing
}
\icmltitlerunning{An Open Model for Real-Time Video Game Playing}


\icmlsetsymbol{equal}{*}
\icmlsetsymbol{p2}{†}

\begin{icmlauthorlist}
\icmlauthor{Yuguang Yue}{p2}
\icmlauthor{Irakli Salia}{p2}
\icmlauthor{Samuel Hunt}{p2}
\icmlauthor{Chris Green}{p2}
\icmlauthor{Wenzhe Shi}{p2}
\icmlauthor{Jonathan J Hunt}{p2}
\end{icmlauthorlist}

\icmlaffiliation{p2}{Player2, USA}

\icmlcorrespondingauthor{Yuguang Yue}{yuguang@elefant.gg}

\icmlkeywords{Machine Learning, Action Policy}

\vskip 0.3in
]



\setcounter{footnote}{2}

\makeatletter
\def\icml@affiliation@mark#1{\textsuperscript{*}}
\makeatother

\printAffiliationsAndNotice{}  

\begin{abstract}
Behavior cloning has seen a resurgence as scaling model and data sizes demonstrate strong performance.
In this work, we introduce an open recipe for training a video game playing foundation model designed for inference in realtime on a consumer GPU. We release all data (8300+ hours of high quality human gameplay), training and inference code, and pretrained checkpoints under an open license. Empirically, we show that our best model achieves performance competitive with human players across a variety of 3D games\footnotemark[1].
We use this recipe to investigate the scaling laws of behavior cloning, with a focus on causal reasoning.
In a controlled toy setting, we first demonstrate that increasing training data and network depth leads to the model learning a more causal policy. We then validate these findings at scale, analyzing models up to 1.2 billion parameters. We observe that the causal improvements seen in the toy domain hold true as model size and training steps increase.
\end{abstract}
\footnotetext[1]{All data, code, model checkpoints and videos of game playing are available from the accompanying website \url{https://elefant-ai.github.io/open-p2p/}.}

\begin{figure*}[t]
    \centering
    \includegraphics[width=\textwidth]{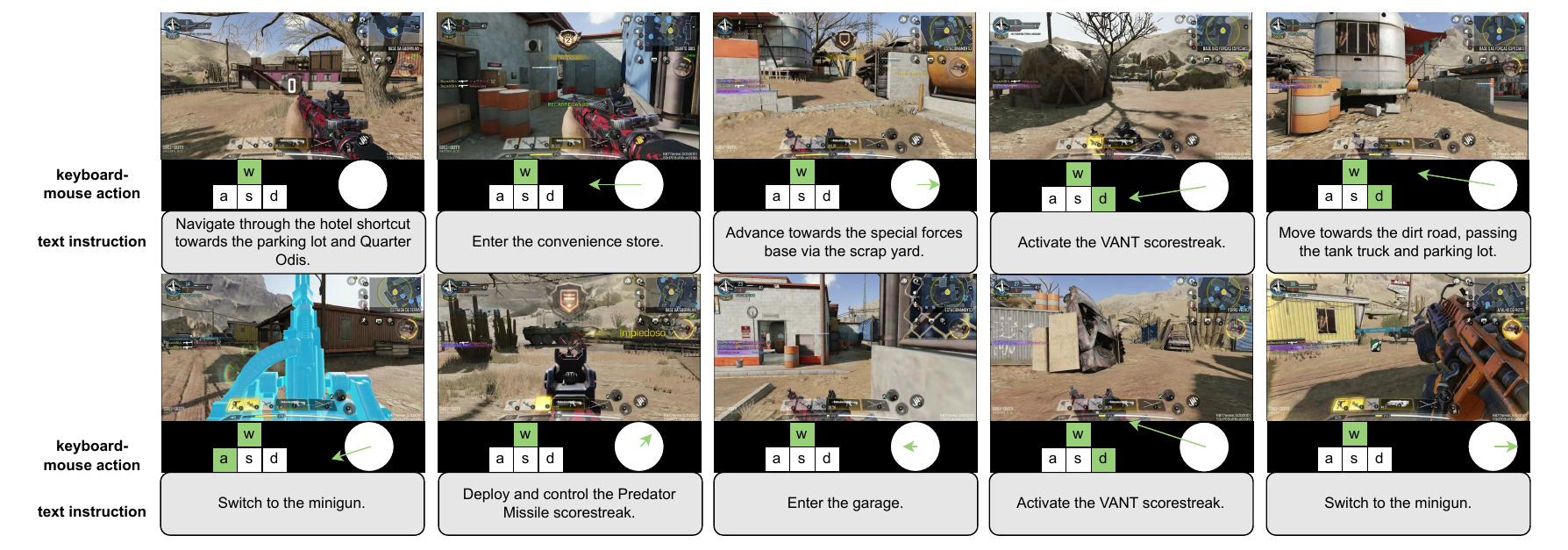}
    \caption{Example gameplay sequence with aligned action and text annotations. For visual clarity, we only show the frames where a text annotation is initialized, keyboard actions are simplified to \texttt{WASD} inputs, and mouse clicks are omitted; The highlighted key means the key is pressed, and the arrow indicates mouse movement in the $x$ and $y$ directions.}
    \label{fig:data-viz}
\end{figure*}

\section{Introduction}
Artificial intelligence (AI) has been applied to game playing since its inception \citep{turing1953digital}. When models are allowed to interact directly with an environment, reinforcement learning has achieved remarkable success, including superhuman performance in complex games \citep{berner2019dota, vinyals2019grandmaster}. However, such systems are typically tailored to a single game, as they rely on carefully engineered training environments and substantial manual design of reward functions, limiting their generality and scalability.

Behavior cloning (BC), by contrast, is a simple and long-standing approach to policy learning that formulates control as supervised learning from state–action pairs \citep{pomerleau1989alvinn, bain1995framework}. Because it learns solely from collected datasets, behavior cloning has the potential to generalize across diverse game environments without requiring environment-specific reward engineering. Nevertheless, BC is known to suffer from two fundamental challenges: distributional shift \citep{ross2011reduction} and causal confusion \citep{de2019causal}, both of which can severely degrade performance.



In this paper, we train a single model capable of playing a range of 3D video games using raw image observations and producing keyboard and mouse actions in real time on a consumer-grade GPU. Our approach leverages BC with a large-scale dataset collected across a diverse set of games. We release all training and inference code, as well as the dataset (an example is shown in Figure~\ref{fig:data-viz}), under open licenses. We demonstrate that the model can play simple games that do not require complex planning with a high level of competence.

We study the scaling laws of BC models under data-constrained regimes by training four model sizes ranging from 150M to 1.2B parameters across five dataset size ranges of 6\%, 12\%, 25\%, 50\%, and 100\% of the full training data. We observe a clear power-law relationship between test loss and dataset size.

Causal confusion occurs due to non-causal correlations in the data, which can result in the policy learning to predict the action using these non-causal correlates. An example, taken from \citet{de2019causal}, is that the policy may learn to apply the brakes when it sees the brake lights since these are highly correlated with braking. Obviously, such a non-causal policy performs poorly.

Using both a simple toy example with a rigorous causality metric and empirical results with an approximate causality score, we show that in our setting, scaling both model and dataset size leads to models that more reliably attend to causal signals. These results suggest that a practical approach to mitigating causality issues in behavior cloning is to scale up model capacity alongside data size and diversity.

\section{Gameplay Dataset}

\subsection{Annotated data}
We collect a large-scale, high-quality dataset of human gameplay spanning a diverse set of popular 3D video games. The complete list of games is provided in Appendix~\ref{Appendix:game-list}. Gameplay is recorded by experienced players who are instructed to capture only active gameplay segments (e.g., excluding lobby or waiting periods). Annotators use a variety of hardware, monitor sizes and resolutions, mouse sensitivities, and gameplay styles, which provide a diverse set of gameplay videos. We do not enforce balanced data distribution across games. 

All gameplay videos are recorded at $20$ frames per second (FPS), following prior work \citep{baker2022video}. For each frame, we capture the raw screen pixels $o_i$ together with the corresponding keyboard and mouse actions $a_i$. A gameplay trajectory is represented as a sequence
\[
[o_1, a_1, o_2, a_2, \ldots, o_T, a_T],
\]


After filtering for quality (see Appendix~\ref{Appendix:quality} for details), the dataset comprises over $8{,}300$ hours of high-quality human gameplay, corresponding to around $600$ million image--action pairs. The distribution of recording hours across games is shown in Appendix~\ref{Appendix:labeled-data}.


\subsubsection{Text-Annotated Data}
To enable text-conditioned policy learning, we augment the gameplay data with text annotations. These annotations provide text instructions that the model is trained to follow over temporal windows ranging from several seconds to a few minutes.

Text annotations were generated retrospectively using a commercial VLM. The VLM was prompted to review gameplay segments and infer plausible instructions that could have guided the human player at specific timestamps. This task is inherently challenging, as it requires inferring the player’s underlying intent from recorded behavior and does not admit a unique mapping.  While VLMs are not currently capable of playing games at a high level \citep{zhang2025videogamebench}, we found that they produce good-quality instructional descriptions when analyzing gameplay videos offline, as shown in Figure~\ref{fig:data-viz}. 

A key challenge in this process is that commercial VLMs typically operate on temporally compressed videos (often around 1~Hz), which is insufficient for text annotation in fast-paced games. To address this limitation, we deliberately prompt the VLM to generate temporally extended, goal-oriented instructions (e.g., ``move toward the skull gate'') rather than instantaneous commands (e.g., ``turn left''). We further design the prompt to suppress repetitive, high-frequency events, such as shooting, by annotating such behaviors only once per contiguous segment. Finally, the VLM is prompted to generate start and end timestamps, at second-level precision, for each instruction, which are then aligned with the original video recordings.

A single unified prompt is used across all games to ensure scalability and avoid per-game customization. The full prompt is provided in Appendix~\ref{Appendix:text-annotation}. 

\begin{figure*}[htbp]
    \centering
    \begin{subfigure}[t]{0.68\linewidth}
        \includegraphics[width=\linewidth]{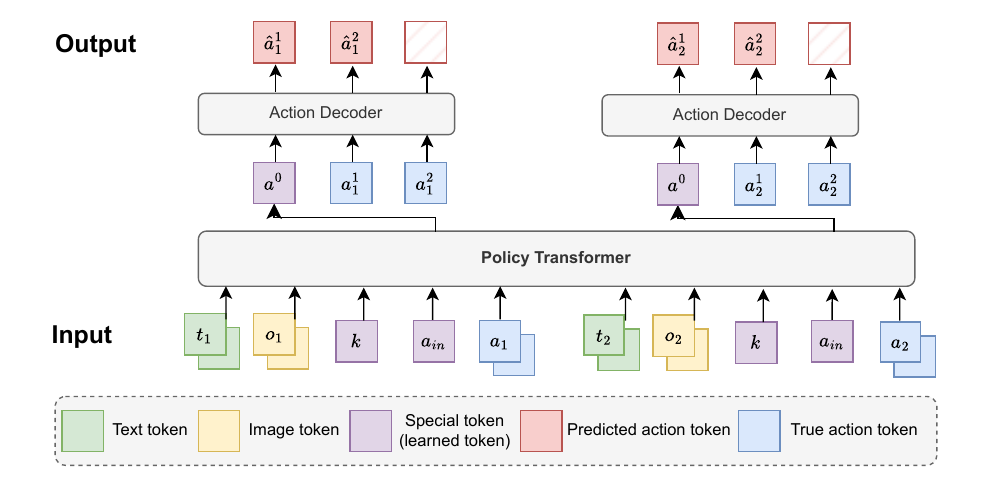}
        \caption{}
        \label{subfig:model}
    \end{subfigure}
    \hfill
    \begin{subfigure}[t]{0.25\linewidth}
        \includegraphics[width=\linewidth]{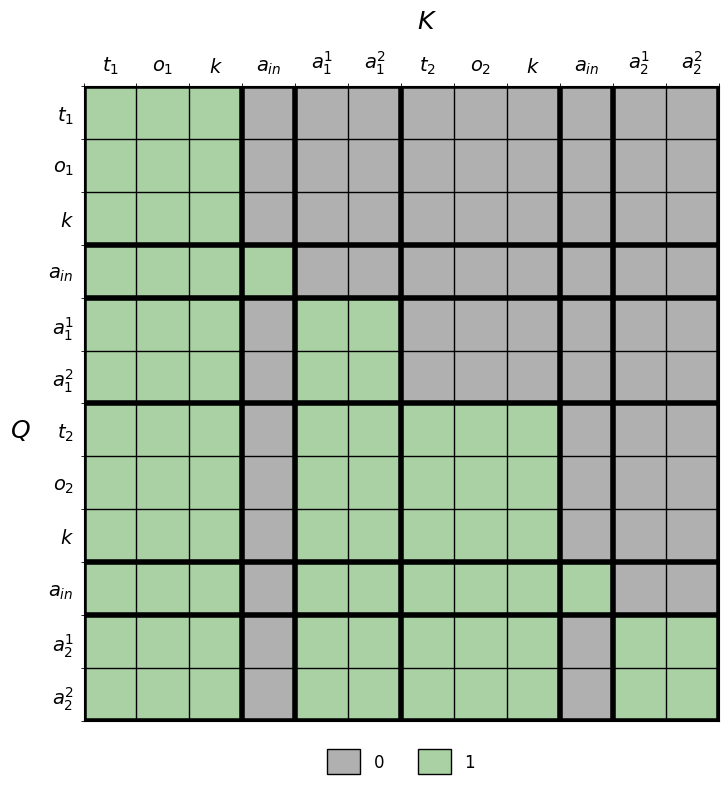}
        \caption{}
        \label{subfig:mask}
    \end{subfigure}
\caption{
    (\subref{subfig:model}) Architecture of P2P. The core policy transformer and action decoder are both decoder-only transformers.
Each timestep begins with a text token $t_i$. Since many frames do not contain a text annotation there is a default text token $t_{null}$ used on these frames. This is followed by image token(s) from video frame $o_{i}$ 
followed by a learnable “reasoning’’ token $k_{i}$ that grants the model extra computation. The policy transformer then outputs a single action prediction token $a_{in}$. We refer to the reasoning token and the action prediction token as special tokens in the architecture. A smaller transformer, the action decoder, then auto-regressively transforms and samples the single action prediction token into the full action space. Then the true action tokens $a_i$ are input so that $a_{in}$ at time $i+1$ can attend to the true action tokens from time $i$.
    (\subref{subfig:mask})
    Attention mask used in our transformer policy (green denotes $1$ and gray $0$). This custom mask ensures the action prediction token $a_{in}$ at time $i$ cannot attend to the ground truth action at time $i$. Note that no other tokens attend to $a_{in}$ to stabilize the training process. 
}
\end{figure*}

\subsubsection{Correction Data}
\label{sec:correction-data}
A common challenge of behavior cloning is distributional shift, where the state distribution encountered during online deployment deviates from that of the training data. Inspired by DAgger \citep{ross2011reduction}, we mitigated this issue by collecting human correction data.

To collect correction trajectories, we deploy a trained policy to interact with the game environment while a human annotator monitors its behavior. When the policy encounters out-of-distribution situations (e.g., becoming stuck or exhibiting degenerate behavior), the annotator temporarily takes control to guide the agent back to a valid state. Control is then returned to the policy. For behavior cloning, we used only the human actions and masked out the loss for actions taken by the agent. These correction trajectories are mixed with the original annotated data during training and constitute less than 1\% of the total annotated dataset.

\subsubsection{Simple Benchmark Environments}
In addition to commercial games, we collected data from two lightweight 3D environments we designed for automated evaluation: \emph{Hovercraft} and \emph{Simple-FPS} (see Appendix~\ref{Appendix:simple-env}). These environments are fully programmatic, allowing precise control over game state and difficulty, which provides a fair and controlled benchmark for comparing model performance (see Section~\ref{sec:eval}).

\subsection{Unlabeled Data}
In addition to annotated gameplay data, we curated a large corpus of unlabeled gameplay videos from public sources \citep{fan2022minedojo, baker2022video}. Details of the unlabeled dataset and its usage are provided in Appendix~\ref{sec:unlabeled}.







\section{Policy Model}\label{sec:policy-model}

We present a multimodal action policy model, which we refer to as \emph{Pixels2Play} (P2P). P2P is a text-conditioned policy that takes visual observations and optional textual instructions as input and outputs low-level keyboard and mouse actions. The model builds upon the transformer-based policy architecture introduced in \citet{yue2025pixels}, but introduces substantial extensions to support text conditioning and to improve online performance by explicitly conditioning the backbone transformer on ground-truth action tokens during training.

A primary design constraint is that the model must operate in real time (20~Hz) on high-end consumer GPUs (e.g.\ NVIDIA RTX~5090), enabling deployment on end-user hardware. To satisfy this requirement, we trained a lightweight, decoder-only transformer backbone from scratch rather than fine-tuning a large pretrained VLM, as explored in prior work \citep{kim2024openvla, tan2025lumine}. This design offers two advantages: (i) it enables a custom image tokenization pipeline that produces a small number of image tokens, allowing the model to attend to longer temporal histories, which is important for games that require long-term memorization; and (ii) the reduced model size and custom architecture ensures fast inference and compatibility with compilation and optimization techniques.

We evaluated several image encoders, including MAGVIT-v2 \citep{yu2023language, luo2024open}, DINOv2 \citep{oquab2023dinov2}, IBQ \citep{shi2025scalable}, and CosmosTokenizer \citep{agarwal2025cosmos}. 
We found that EfficientNet \citep{tan2019efficientnet}, following \citet{pearce2022counter}, provides the best trade-off between representation quality and computational efficiency. Specifically, we use the first six layers of EfficientNet, followed by a linear projection into a small set of visual tokens (typically $N_i \in \{1,\dots,4\}$). Unfreezing the image encoder during training consistently improves performance compared to freezing it (Appendix~\ref{Appendix:frozen}). The model operates directly on the resized raw pixel inputs (192 pixels $\times$ 192 pixels).

Much prior work focused on single-game agents and adopted reduced action spaces that can be modeled as a single categorical variable \citep{baker2022video, pearce2022counter}. In contrast, our setting requires a unified policy capable of operating across many games, necessitating a much larger action space comprising the full keyboard and mouse input domain. We allowed up to four simultaneous key presses and two concurrent mouse actions, making treating the action as a single one-hot combinatorial approach impractical.

To address this, we modeled actions autoregressively. We used $N_a=8$ tokens for each action: $4$ keyboard tokens, $2$ mouse tokens (x, y movement), and $2$ mouse button tokens. To avoid increasing the token count of the main transformer, we introduced a lightweight \emph{action decoder}. The backbone policy transformer outputs a single latent action token $a_{\text{in}}$, which is then decoded autoregressively by the action decoder into the full action specification (Figure~\ref{subfig:model}). This design allows the policy transformer to perform a single forward pass per timestep during inference, yielding approximately a $5\times$ speedup in real-time execution compared to directly predicting all action tokens.

For text embeddings, we use EmbeddingGemma \citep{vera2025embeddinggemma} to encode textual inputs and apply mean pooling over token embeddings to obtain a single text representation. Text embeddings are precomputed, and the text tokenizer is frozen during training.

At each timestep, the policy transformer consumes image tokens ($o_i$), a text-conditioning token ($t_i$), ground-truth action tokens ($a_i$), and a single action prediction token ($a_{\text{in}}$). We further introduce an optional ``thinking'' token $k$, which provides the model with an additional reasoning step prior to action prediction.  We refer to the reasoning token and the action prediction token as special tokens in Figure~\ref{subfig:model}.
The resulting token count per timestep is $N_i + N_a + 3$. Each token is augmented with a learned type embedding indicating its role (image, text, reasoning, ground-truth action, or action prediction). Rotary positional embeddings \citep{su2024roformer} are applied at every transformer layer. During inference, we employ key--value caching with sliding-window attention to bound memory usage. 

Because ground-truth action tokens are provided as input during training, we design a custom causal attention mask to prevent information leakage. The action prediction token $a_{\text{in}}$ is prohibited from attending to ground-truth action tokens at the same timestep, ensuring causality. Other tokens (image, text, reasoning, and ground-truth actions) may attend to each other within the same timestep, but are restricted from attending to $a_{\text{in}}$ tokens from previous timesteps to avoid training instability (Figure~\ref{subfig:mask}).

Behavior cloning often suffers from causal confusion \citep{de2019causal}. One particular case of this, which becomes worse at higher frequencies, is that the model learns to copy previous actions instead of attending to visual inputs \citep{wen2020fighting}. Although this issue can be avoided by excluding ground-truth actions from the policy inputs, we find that conditioning the policy on ground-truth action tokens produces more human-like behavior than conditioning on visual observations alone. In particular, the policy learns to sustain actions over multiple frames before switching, more closely resembling human gameplay dynamics. Qualitative differences in behavior are illustrated at \url{https://elefant-ai.github.io/open-p2p/}.

We observe pronounced causal confusion when directly predicting action tokens without an action decoder and when we only had a small amount of training data. In contrast, introducing an action decoder and scaling the dataset size substantially mitigates this issue. A detailed analysis is provided in Section~\ref{sec:causality}.

\subsubsection{Leveraging unlabeled data}

Since unlabeled data from publicly available resources are far more abundant than annotated data, it is desirable to leverage such unlabeled data for training. Details on leveraging unlabeled data, including the training procedure and some preliminary experimental results, are provided in Appendix~\ref{Appendix:pretrain}.

\subsection{Mitigating the Training--Inference Gap}
Early experiments revealed a substantial performance gap between offline evaluation and online deployment. We traced this gap to discrepancies between training and inference inputs arising from video recording, compression, and resizing. For practical latency and storage considerations, videos undergo two lossy processing steps during data collection: (1) compression and upload on the annotator side, and (2) compressing and resizing to $192 \times 192$ for model processing. During inference resizing occurs but no compression takes place.

\begin{figure}[ht]
  \centering
  \begin{subfigure}[b]{0.5\textwidth}
    \includegraphics[width=\linewidth]{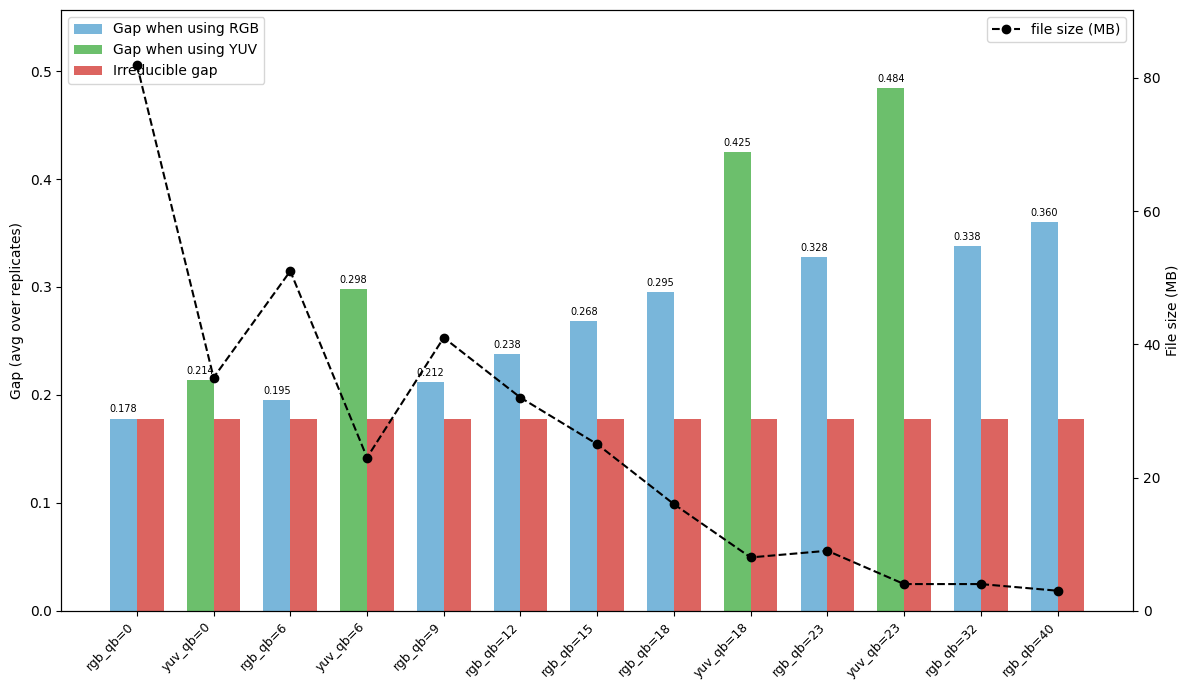}
    \caption{}
    \label{fig:no-augmentation}
  \end{subfigure}
  \hfill
  \begin{subfigure}[b]{0.5\textwidth}
    \includegraphics[width=\linewidth]{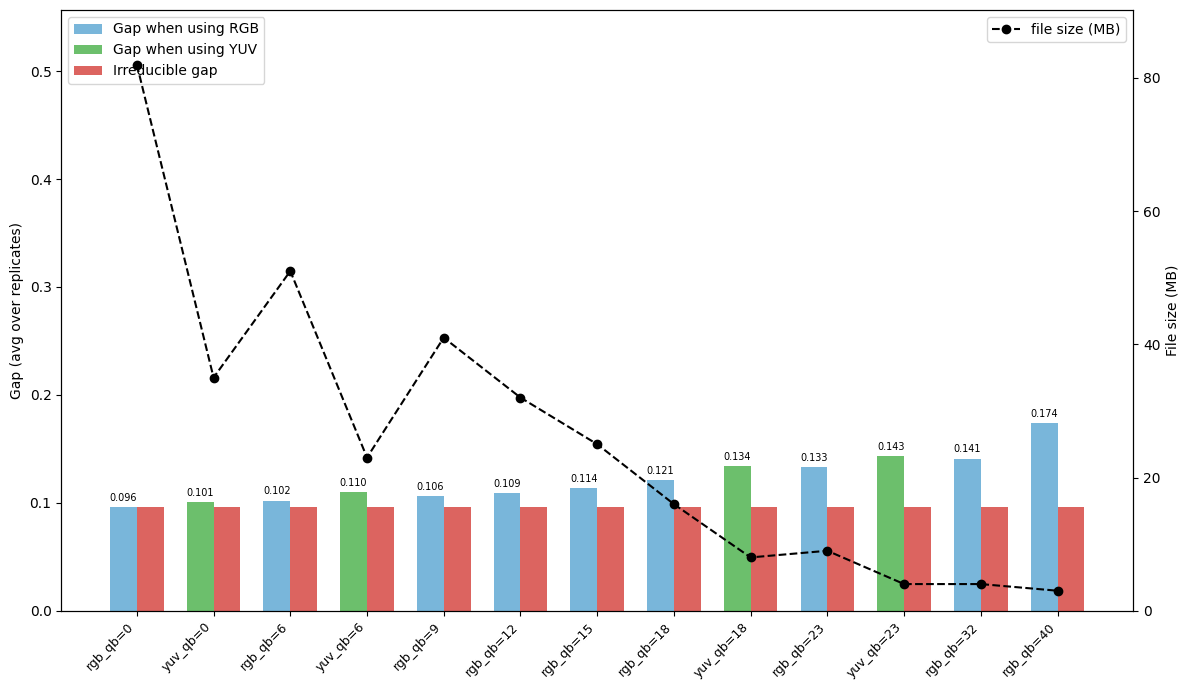}
    \caption{}
    \label{fig:augmentation}    
  \end{subfigure}
\caption{(\subref{fig:no-augmentation}) Gap induced by video compression without data augmentation. We measured this gap by comparing model outputs on raw frames (inference) and resized frames (training). An irreducible gap arises from lossy video compression during data collection. The gap was smaller with RGB than YUV encoding and increases as compression quality degrades (lower file size). The x-axis \texttt{qp} denotes the quantization parameter, where larger values indicate lower quality. (\subref{fig:augmentation}) Data augmentation  mitigates this gap when reasonable compression quality is used.}
\end{figure}

We observed a substantial divergence in model outputs when using uncompressed raw frames versus resized training frames, which leads to deceptively strong training metrics but poor online performance. We measured this gap by collecting a small number of uncompressed videos. We then compressed the videos using differing compression options and compared the trained model probabilities on the uncompressed versus compressed output.
Although compression at the annotation stage is unavoidable (which causes the irreducible gap), we find that the choice of color space during resizing plays a critical role: RGB encoding yields a smaller training–inference gap than YUV encoding (Figure~\ref{fig:no-augmentation}). Unfortunately, NVIDIA hardware encoders support only the YUV color space; therefore, we adopt a mixture of QP values ranges from $6$ to $18$ to balance encoding speed and encoding quality.

We also found that two different resizing functions were used between inference and training code paths that, while visually indistinguishable, contributed to the training–inference gap. The training code used a PyTorch function, while inference used a Rust function. We modified the code to ensure a bitwise identical resizing function was used for both training data and inference, and this mitigated the issue.

Data augmentation also substantially reduces the training–inference gap. We applied mild spatial transformations, color perturbations, Planckian jitter, ISO noise, random gaussian or motion blur, sharpening, and translation during training.
As shown in Figure~\ref{fig:augmentation}, these augmentations significantly reduced the discrepancy and improved online performance. Consequently, all experiments in this work employ data augmentation. We believe further improvements to reduce the training-inference gap via targeted data augmentation are an interesting area for future work. The detailed parameters are shown in Table~\ref{tab:data-augmentation}.

Finally, discretization of the mouse action space can lead to overly aggressive or overly conservative mouse movements. Following \citet{pearce2022counter}, we discretized mouse actions using quantile-based bins, which provide fine resolution near zero but coarse resolution in the tails. To improve robustness, we fit truncated normal distributions to the x and y axes of the mouse action space using undiscretized training data. At inference time, the policy’s predicted discrete mouse action defines the upper and lower bounds, from which we sample the final mouse action using the fitted truncated normal distribution.
The mean and standard deviation estimated from the mouse movements data are $\hat{\mu}_x = 0, \hat{\mu}_y = 0, \hat{\sigma}_x=96, \hat{\sigma}_y=22$. Empirically, this approach yields smoother control and better online performance.

\section{Evaluation}
\label{sec:eval}
We trained policy models at four parameter scales (150M, 300M, 600M, and 1.2B) and compared their performances. More technical details of the training parameters are in Appendix~\ref{appendix:training}.

We evaluated our models using (i) the scores from controlled programmatic environments, (ii) human preference evaluations on real games, and (iii) quantitative analyses of scaling behavior on test loss\footnotetext[2]{Perplexity of the keyboard was used as test loss because some games do not require mouse actions so mouse perplexity was more noisy.} and causality. Although several game benchmarks exist \citep{zhang2025videogamebench, park2025orak, xu2025probe, tomilin2023coom}, they primarily focus on text-heavy 2D games or sandbox-style 3D environments, which differ substantially from the real-time, first-person gameplay data used in our training setup. 

Textual input is used exclusively for the instruction-following evaluation and is not provided in any other evaluation setting.

Camera settings can significantly impact performance when playing games in real time, as overly high sensitivity makes it difficult for the model to adjust to small movements. Detailed camera configurations are provided in Appendix~\ref{Appendix:camera-setup}.

\subsection{Simple Programmatic Environment}\label{sec:simple-env}
We first evaluate our models in two programmatic environments: \emph{Hovercraft} and \emph{Simple-FPS}. These environments were implemented by us in Godot \citep{holfeld2023relevance}, allowing full control over map layouts and difficulty settings, which enables fair and reproducible comparisons across model variants.

For Hovercraft, we measure the time (in seconds) required for the agent to complete a full loop. For Simple-FPS, we report the number of hits on the enemy minus the number of hits received. We also report end-to-end inference latency measured on a single RTX 5090 GPU. We compared models of varying sizes trained on the full labeled dataset.
Each model was evaluated $16$ times in the same environment, and we report the mean and standard deviation. The results are summarized in Table~\ref{tab:godot-env}. Overall, the largest model achieves the highest mean score with the lowest variance.

\begin{table}[t]
\centering
\small
\rowcolors{2}{gray!10}{white}
\begin{tabular}{lccc}
\toprule
\textbf{Model Size} &
\textbf{Hovercraft $\downarrow$} &
\textbf{Simple-FPS $\uparrow$} &
\textbf{FPS $\uparrow$} \\
\midrule
150M & $41.1\pm 8.8$ & $25.3 \pm 6.5$ & \textbf{80} \\
300M & $41.7\pm 8.2$  & $25.1\pm 5.6$ & 64 \\
600M & $38.1 \pm 4.2$ & $26.5\pm 10.4$ & 62 \\
1.2B & $36.8\pm 2.9$  & $26.6\pm 3.7$ & 40 \\
\bottomrule
\end{tabular}
\caption{Performance on Godot-based programmatic environments across model sizes. For Hovercraft, the score is the time (in seconds) required to complete a loop; for Simple-FPS, the score is the number of enemy hits minus the number of hits received. Each model is evaluated $16$ times per environment, and we report the mean $\pm$ standard deviation. FPS denotes the achievable inference throughput (frames per second) on an RTX 5090 GPU.}
\label{tab:godot-env}
\vspace{-20pt}
\end{table}

\subsection{Human Evaluation in Real Environments}
\label{sec:unlabelled}

Because our policy is designed for human-like real-time interaction in real video games, it is challenging to evaluate. We used human evaluation on gameplay videos generated by the models. We evaluate performance across four games: two single-player titles (DOOM and Quake from the Steam platform) and two multiplayer games (Roblox \emph{Be-a-Shark} and Roblox \emph{Hypershot}). For DOOM and Quake, we manually divide each game into three checkpoints. The model is initialized from each checkpoint and run for one to two minutes or until it reaches the subsequent checkpoint. To reduce evaluation variance, we run the model three times for each checkpoint in DOOM and Quake. Together with \emph{Be-a-Shark} and \emph{Hypershot}, this results in a total of $20$ evaluation videos for each model.

Human evaluators assessed model quality by counting the occurrences of the following issues during gameplay: (1) colliding with walls; (2) shooting into the air; (3) missing targets (including items or enemies); (4) exhibiting non–human-like behavior (e.g., repeating loops or moving backward); (5) remaining idle; and (6) camera shaking or jittering. For each gameplay video, evaluators recorded the number of occurrences of each issue and normalized the counts by the video length. Lower values indicate better performance. Evaluators were blinded to the model that generated the video.

\begin{figure}[ht]
  \centering
  \begin{subfigure}[b]{0.5\textwidth}
    \includegraphics[width=\linewidth]{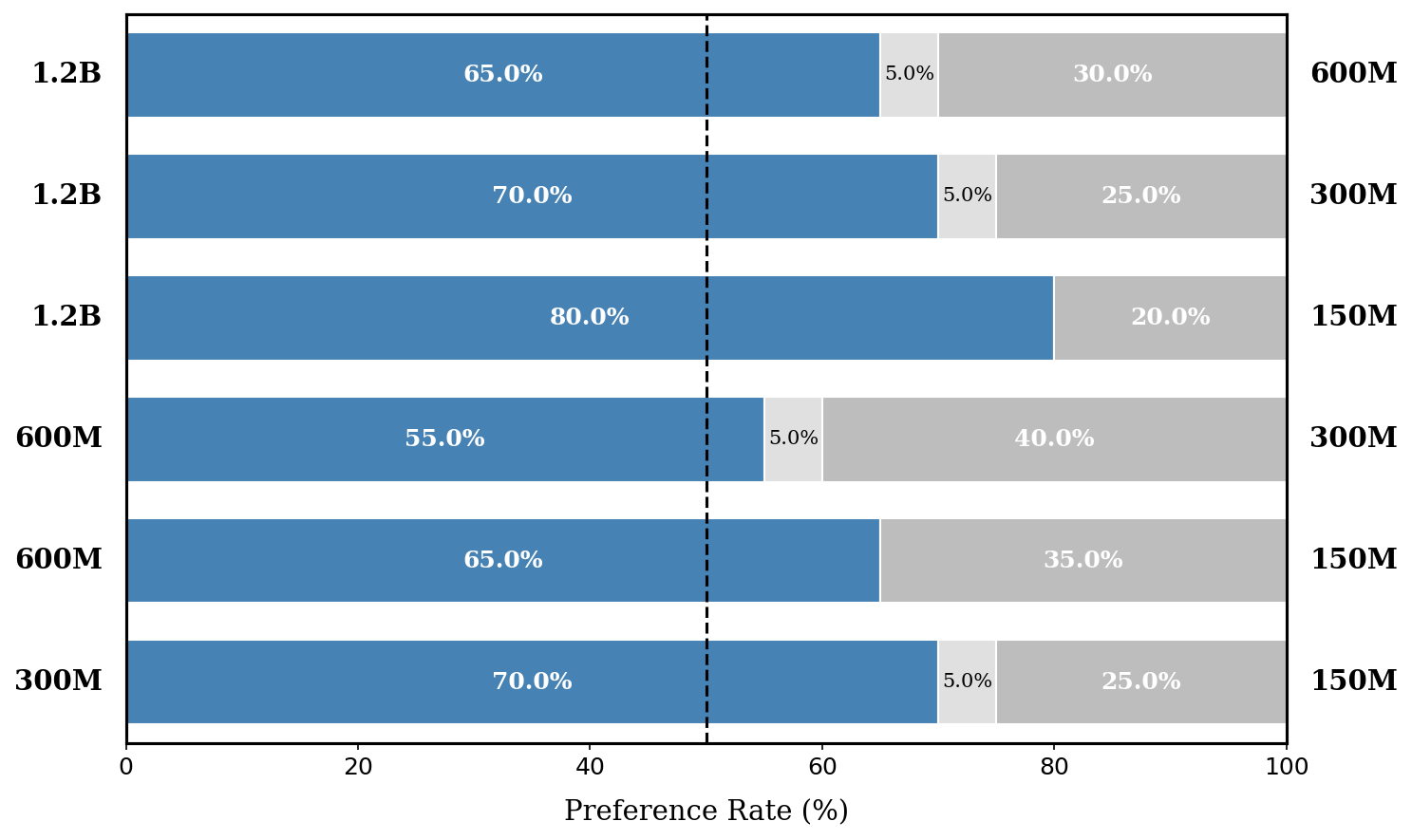}
    \subcaption[]{}
    \label{fig:model-comparison}    
  \end{subfigure}
  \hfill
  \begin{subfigure}[b]{0.5\textwidth}
    \includegraphics[width=\linewidth]{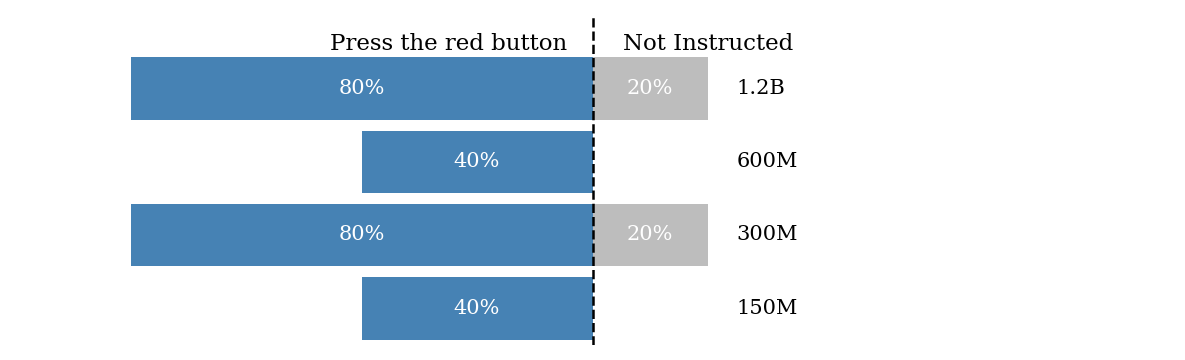}
    \subcaption[]{}
    \label{fig:instruction-following}    
  \end{subfigure}
  \caption{
  (\subref{fig:model-comparison}) Human preference comparisons across model sizes. Blue and gray bars indicate preferences for each model, while the centered numbers denote the percentage of ties. Each model is evaluated on 20 game checkpoints, with gameplay trajectories recorded.
  (\subref{fig:instruction-following}) Instruction-following comparison. Each model is evaluated from the same maze checkpoint with and without an instruction (``press the red button''), with five runs per condition. We report the success rate of completing the maze.
  }
  \vspace{-18pt}
\end{figure}

Figure~\ref{fig:model-comparison} presents quantitative preference comparisons across model sizes, with the underlying rubric scores reported in Appendix~\ref{Appendix:human-eval}. Larger models are consistently preferred over smaller ones, in agreement with the numerical results in Section~\ref{sec:simple-env}.

\subsubsection{Instruction-following} 
We evaluated the model’s ability to follow text instructions. We performed this evaluation in the Quake environment, where experiments can reliably start from the same checkpoint. We selected a maze scenario in which the player must press three red buttons on the wall to unlock a door (see Appendix~\ref{Appendix:text-instruction} for details).

Without textual input, the model often fails to press all three buttons. This failure mode is expected, as the policy primarily imitates expert trajectories, and the action of not pressing a button introduces only subtle deviations in behavior unless explicitly emphasized by instruction. We evaluated all models on this maze over five runs and compared their success rates. As shown in Figure~\ref{fig:instruction-following}, providing the text instruction “press the red button” substantially increases the success rate compared to the no-text baseline for all the models, demonstrating that the model actively conditions its behavior on text input.

We note that the model can currently follow only simple instructions that are similar to those seen during training, due to the limited quantity and diversity of textual instructions in the training data. We expect this limitation to be alleviated by scaling up text instructions in both diversity and volume.

\begin{figure}
    \centering
    \includegraphics[width=1.0\linewidth]{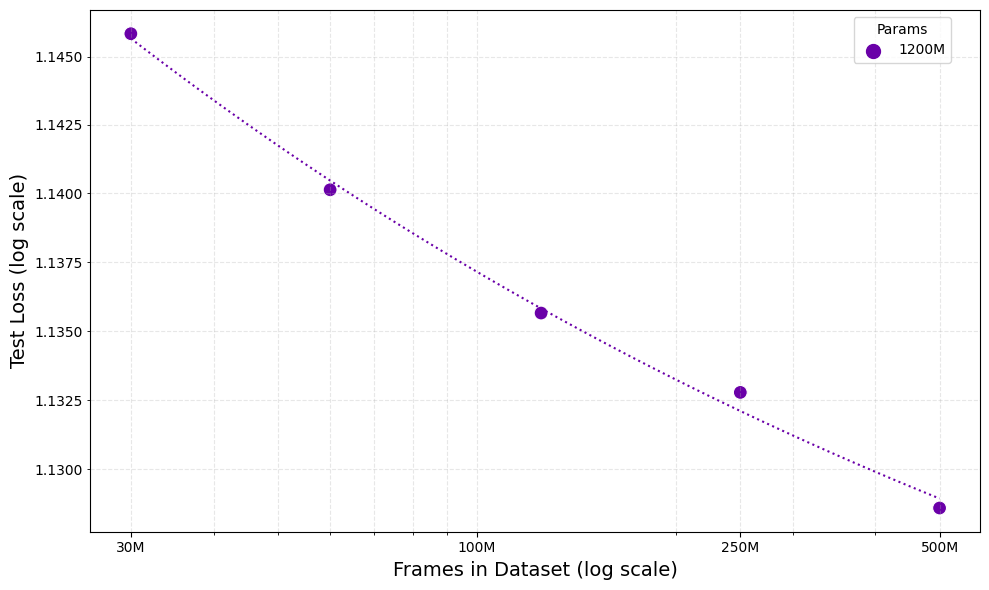}
    \caption{Lowest test loss versus dataset size for the 1.2B model. As might be expected, we find the test loss fits a power-law curve closely.}
    \label{fig:scaling-law-1-2b}
    \label{subfig:scaling-law-12b}
\end{figure}

\section{Causality and scaling laws}
We used our working recipe for model training and architecture to investigate the relationship between model and dataset size and both the test loss and the causal behavior of the model. We focused on the data constrained regime, which is becoming an increasing focus in other modalities \citep{muennighoff2025dataconstrained, kim2025pre, pearce2024scaling}. In this regime, we train each model/data size for multiple epochs until overfitting or a lack of further improvement is observed.

 We used a static subset of approximately 500M frames (about 7,000 hours of gameplay). Models are trained on fractions of this dataset \((100\%, 50\%, 25\%, 12\%, 6\%)\) across four parameter scales \((1.2\text{B}, 600\text{M}, 300\text{M}, 150\text{M})\). 


Following \citet{kaplan2020scaling}, we fitted the empirical scaling relationship
\[
L(D) = L_{\infty} + \left(\frac{D_c}{D}\right)^{\alpha},
\]
where $L_{\infty}$ denotes the irreducible loss, $D$ is the number of training frames, and $D_c$ and $\alpha$ are fitted constants.

For the 1.2B model, we estimate $L_{\infty} = 1.111$, $\alpha = 0.2336$, and $D_c = 17$, as shown in Figure~\ref{fig:scaling-law-1-2b}. Scaling curves for all model sizes are reported in Appendix~\ref{Appendix:scaling-law}. 

In Appendix~\ref{Appendix:scaling-law} Figure~\ref{fig:scaling-law-appendix}, we provide the full test loss trajectories as a function of training steps for all combinations of model scale and dataset size.
These results show that, in general, larger models achieve lower test loss, and increasing the amount of training data consistently reduces test loss. Moreover, larger models benefit more from additional data in data-abundant regimes (e.g., when using 50\% or 100\% of the training data).


\begin{figure}[htbp]
    \centering
    \begin{subfigure}{0.23\textwidth}
        \includegraphics[width=\linewidth]{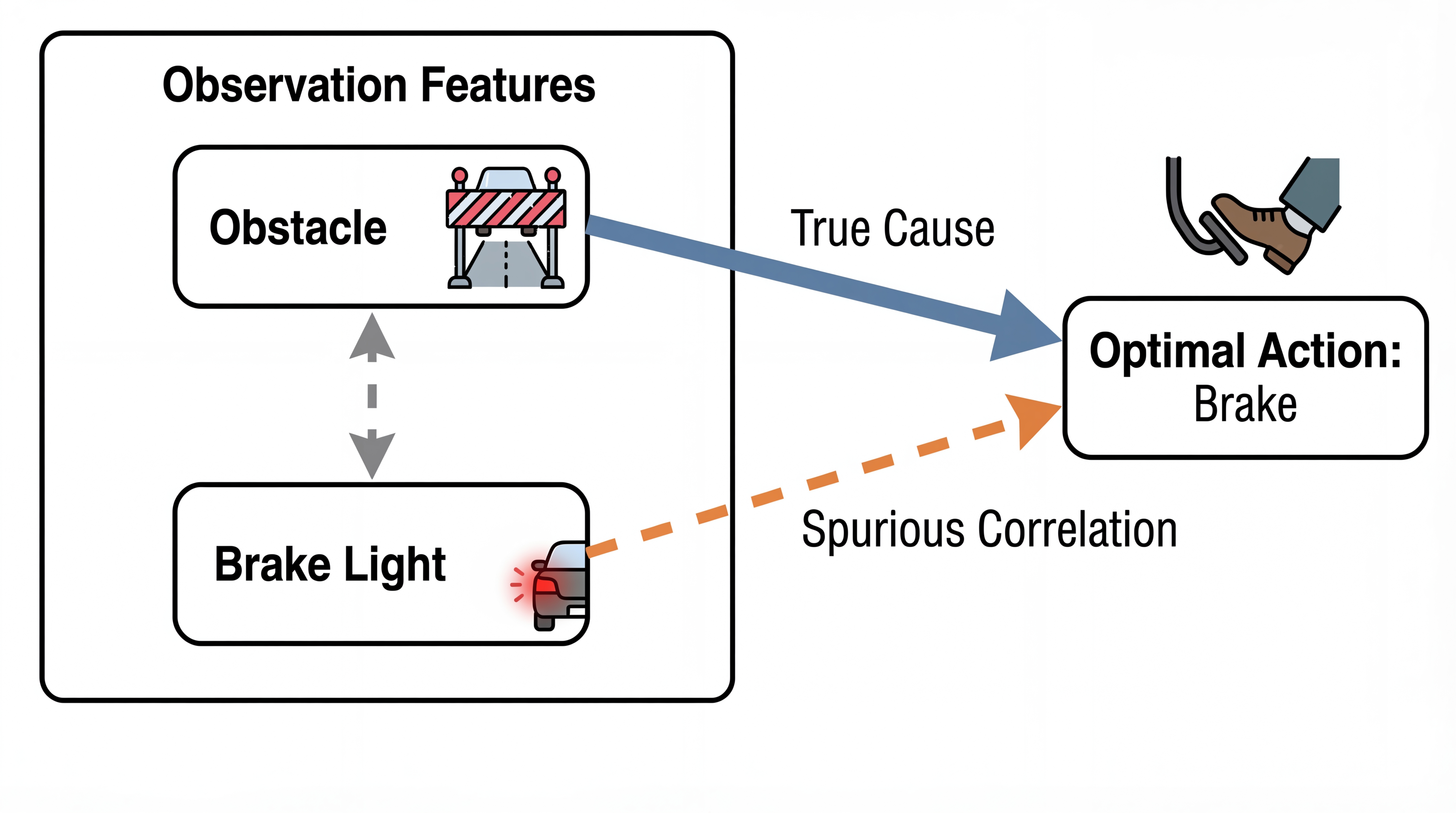}
        \subcaption[]{}
        \label{subfig:toyproblemenv}
    \end{subfigure}
    \begin{subfigure}{0.23\textwidth}
        \includegraphics[width=\linewidth]{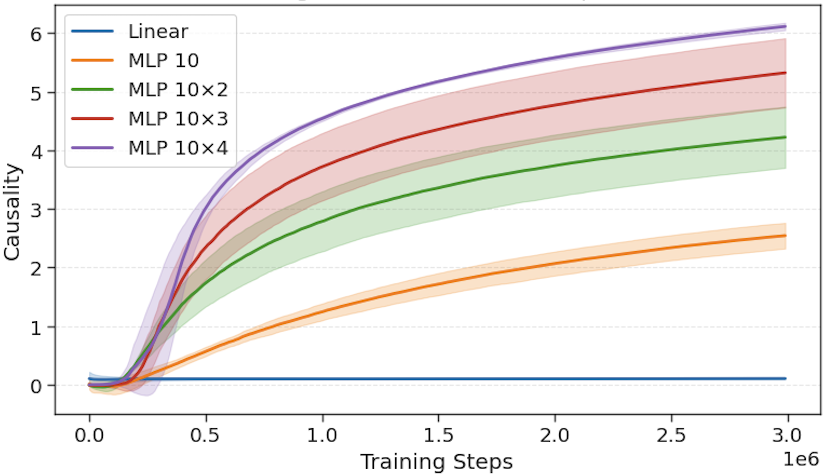}
        \subcaption[]{}
        \label{subfig:toyresults}
    \end{subfigure}
    \caption{
    (\subref{subfig:toyproblemenv}) A toy environment we used to investigate causality in behavior cloning. The observation contains both a causally informative feature (is an obstacle present) and a correlated but non-causal feature (is the brake light on from the previous frame). (\subref{subfig:toyresults})
    We find that increasing the depth of the network improves the speed of learning a causally correct solution. We also find that for all non-linear networks, an approximately causality correct solution is found using SGD. Despite the fact that a optimal linear policy exists, we find that SGD makes no progress towards learning a solution with a randomly initialized linear network.
    }
    \label{fig:toy}
    \vspace{-15pt}
\end{figure}

\subsection{Causality Analysis}
\label{sec:causality}
\subsubsection{Toy problem}

\label{sec:toy}
Before presenting the results on the large behavior cloning models, we first construct a simple environment to better understand the relationship between causality and network depth. As we will show, the results demonstrate surprising commonality with the scaled up models.

In the toy environment, the observation consists of two binary features (Figure \ref{subfig:toyproblemenv}) and three distractor features of random noise. The action is a single binary choice (brake, don't brake). The binary features indicate when an obstacle is present and whether the brake was applied in the previous step (i.e.\ brake light). The optimal policy is to brake whenever an obstacle is present and to ignore all other components of the observation. However, because obstacles persist for multiple frames, there is a strong correlation between the brake light feature and braking.

We train simple neural networks consisting of a linear (with bias) network and multilayer perceptrons (MLPs) of differing depths with ReLU non-linearity. All networks have a final sigmoid, so that the output can be interpreted as $p(a|s)$. All networks are trained using stochastic gradient descent (SGD), a batch size of 1, and a fixed learning rate. The training data consists only of optimal behavior.

We evaluate the causality of the learned policies as $c = \log p(a=1 | s_{o}) - \log p(a=1 | s_{b})$ where $s_{o}$ is an observation with an obstacle and no brake light, and $s_{b}$ is an observation with no obstacle but the brake light on. The optimal causal policy should brake for the obstacle but not for the brake light so $c_{optimal}=\infty$. Note that this measure is logarithmic.

The causally optimal policy can be trivially represented with a linear policy. Despite this, we find that training a randomly initialized linear policy using SGD results in no progress towards learning a causal policy. Adding non-linearity is necessary for the SGD optimization to make progress towards a causally correct solution. We find that increasing the number of layers improves the speed of learning a causally optimal solution (Figure \ref{subfig:toyresults}).

\subsection{Scaled Causality}

In practical settings, such as video game playing, it is much more difficult to have an explicit measure of the causal correctness of a model. We therefore focus on one particular pain-point: in action-conditioned models, predictions can be influenced by two sources: the visual input (frame sequence) and the ground-truth action history. While effective decision-making should largely be causally driven by image observations—analogous to human perception—models often exhibit a ``lazy'' behavior, relying disproportionately on the statistical priors of the action sequence rather than the visual evidence.

To quantify this tendency, we propose a \emph{causality score} that measures the model’s reliance on the input frame sequence. A higher score indicates a stronger causal dependence of the model’s output on frame input. We analyze this score across different model sizes and dataset scales as a function of the number of training frames. Our results show that, as in the toy example, larger, deeper models generally achieve higher causality scores, and that causality increases with additional training. Furthermore, increasing the number of unique training frames leads to higher causality scores, except in extremely data-limited regimes.

We compute the causality score by measuring changes in the model’s output distribution using the KL divergence between predictions produced from the original input sequence and from a perturbed version of the same sequence. Formally,
\[
\text{score}(f) = 
\sum_{b =1} ^{B} 
\sum_{c=1}^{C} 
\mathrm{KL}\big(f(o_{b,c}, a_{b,c})|| f(\tilde{o}_{b,c}, a_{b,c})\big),
\]
where \(f\) denotes the policy model, \(o\) is the original image sequence, $a$ is the original action sequence, \(\tilde{o}\) is the perturbed image sequence, $B$ is the batch size, and \(C\) is the number of evenly sized temporal chunks. 

Given an input sequence, we partition it into \(C\) equal-length chunks and randomly perturb individual image frames with a probability \(p\). The model output at the end of each chunk is then compared between the original and perturbed inputs. Only image frames are perturbed; the action sequence remains unchanged. To ensure that perturbations remain semantically meaningful rather than degenerate noise, each perturbed frame is replaced with a frame drawn from a different scene within the same game. This is implemented by swapping frames within the batch, preserving game-level context while inducing perturbations. We use $C=10, p=0.5, B=32$ for all evaluations. 

\begin{figure}
    \centering
    \includegraphics[width=1.0\linewidth]{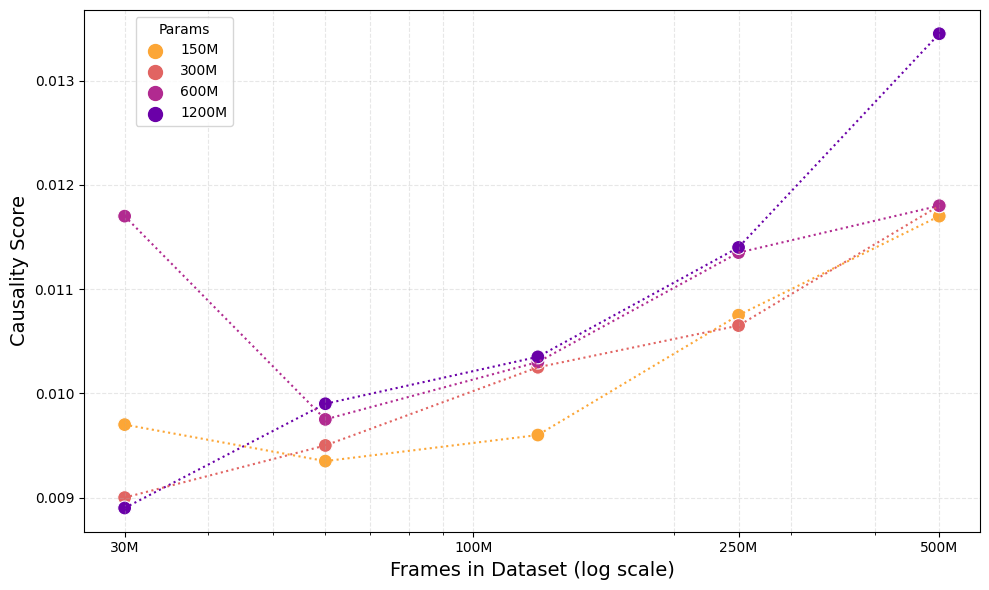}
    \caption{Causality score as a function of dataset size and model size. Except in the low-data regime (30M), the causality score generally increases with larger models and larger training datasets.}
    \label{fig:causality_vs_dataset}
    \label{subfig:causality_vs_dataset}
    \vspace{-15pt}
\end{figure}

For fair comparison across models, we compute causality scores using the checkpoint with the \textit{lowest} test loss for each model. As shown in Figure~\ref{fig:causality_vs_dataset}, causality scores increase with both model size and dataset size, except in the extremely data-limited regime (e.g., $30$M frames). The standard deviation of the score is on the order of $10^{-4}$ and does not affect the relative ranking of models.

We additionally report causality scores as a function of training steps in Appendix~\ref{Appendix:causality}, where we observe a consistent increase throughout training. Notably, causality scores continue to rise even in overfitting regimes, suggesting that both causality scores and test loss should be considered jointly for interpretation.

\section{Related Work}

A large body of prior work studied vision language action (VLA) models for robotic control. Many approaches fine-tune pretrained vision--language models (VLMs) to generate robot actions \citep{driess2025knowledge, kim2024openvla, pertsch2025fast, intelligence2504pi0, zitkovich2023rt, wang2025vla, chen2025combatvla, zhou2025chatvla, shukor2025smolvla, bjorck2025gr00t}, while others train VLA models from scratch using robot-centric video and instruction data \citep{brohan2022rt, zheng2025flare}. These methods have demonstrated strong performance in both simulated and real-world robotic tasks.


A parallel line of work focuses on learning action policies in video game environments. Many studies trained agents within a single environment or testbed, such as Minecraft \citep{li2025optimus, hafner2025training, fan2022minedojo, baker2022video, lifshitz2023steve, wang2024omnijarvis} or Counter-Strike: Global Offensive \citep{pearce2022counter, durst2024learning} and Genshin Impact \citep{tan2025lumine}. Other approaches leverage large language models (LLMs) or vision language models (VLMs) to guide or improve gameplay through planning, code generation, or tool use \citep{yang2024octopus, li2025optimus, wang2023voyager, ma2023eureka}. 

In multi-game settings, \citet{reed2022generalist} demonstrated a generalist agent capable of playing a variety of Atari games, while \citet{wiedemer2025video} leveraged VLMs for zero-shot understanding and control. More recent works train general action models across diverse video game environments \citep{raad2024scaling, bolton2025sima, wang2025game}. However, these systems do not release code or datasets, making reproduction and further research challenging for the broader community.

A contemporaneous piece of work \citep{magne2025nitrogen} also explored training a single model across multiple video games, primarily focusing on controller-based console games (e.g., Xbox). In contrast, our work targets keyboard--mouse based PC games, which involve different interaction modalities and higher-frequency real-time control, making our setting complementary to prior efforts.

\section{Conclusion}
In this work, we present a large-scale, high-quality video dataset annotated with both actions and text, and introduce a real-time policy model designed to leverage this data for online gameplay in real-world environments. We discuss key technical considerations for effectively training such models and evaluate their performance through both qualitative and quantitative analyses. 

Furthermore, we study how model performance scales with dataset size and model capacity, examining both scaling laws and causal behavior. Our results show that larger models achieve lower test loss and higher causality scores in data-abundant regimes. This suggests that one approach to issues of causality in behavior cloning may be simply scaling both model size and dataset size and diversity.

\bibliography{paper}
\bibliographystyle{icml2025}


\newpage
\appendix
\onecolumn
\section{Dataset}
\subsection{Game List}\label{Appendix:game-list}
We list all the games we collected in the annotated dataset:
\begin{itemize}
\begin{multicols}{4}
    \item Roblox: Blade Ball
    \item Roblox: Death Ball
    \item Roblox: Be a Shark
    \item Roblox: Be a Snake
    \item Roblox: Be a Tornado
    \item Roblox: Natural Disaster Survival
    \item Roblox: Rivals
    \item Roblox: Slap Battle
    \item Roblox: A Dusty Trip
    \item Roblox: Hypershot
    \item Roblox: Evade
    \item Roblox: Murderers vs.\ Sheriffs
    \item Msdos: Quake
    \item Msdos: Need for Speed
    \item Msdos: DOOM
    \item Msdos: DOOM II
    \item DOOM Eternal
    \item Eval: Hovercraft
    \item Eval: Simple FPS
    \item Left 4 Dead 2
    \item Call of Duty Mobile
    \item Call of Duty: Black Ops II Zombies
    \item Call of Duty: Black Ops III Zombies
    \item Goat Simulator
    \item Goat Simulator 3
    \item Helldivers 2
    \item Need for Speed: Hot Pursuit
    \item Need for Speed: Most Wanted
    \item Need for Speed: Carbon
    \item Need for Speed: Underground 2
    \item Fortnite
    \item House Flipper
    \item House Flipper 2
    \item PowerWash
    \item Euro Truck Simulator 2
    \item Warhammer: Vermintide
    \item Warhammer: Vermintide 2
    \item Saints Row: The Third
    \item Saints Row IV
    \item Resident Evil 5: Mercenaries
    \item Resident Evil Revelations 2: Raid
    \item Grand Theft Auto V
    \item Grand Theft Auto: San Andreas
    \item Minecraft
\end{multicols}
\end{itemize}

\subsection{Labeled Data}\label{Appendix:labeled-data}
\begin{figure}[H]
    \includegraphics[width=\textwidth]{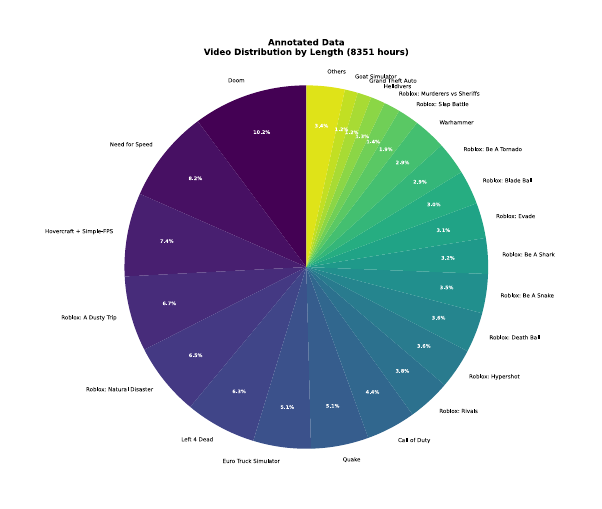}
    \caption{Distribution of games in the annotated dataset.}
    \label{fig:dataset-distribution-comparison}
\end{figure}

The labeled data distribution regarding game types can be found at Figure~\ref{fig:dataset-distribution-comparison}

\subsection{Quality filter}\label{Appendix:quality}

Each recorded video undergoes a two-stage quality assurance process. First, we apply automated filtering based on the following criteria:
(1) no single key is held for more than $60\%$ of the total frames to avoid hardware or recording artifacts;
(2) no more than six keys are pressed simultaneously at any time;
(3) a minimum level of interaction is maintained, ensuring that actions are taken throughout the sequence;
(4) a lightweight action policy model (described in Section~\ref{sec:policy-model}) is evaluated on the full trajectory, and videos with likelihoods below a predefined threshold are flagged.

All videos flagged by these checks are reviewed by human inspectors. In addition, we randomly sampled videos for manual inspection using an adaptive strategy based on each annotator’s prior contributions to a given game: annotators with fewer accepted, recorded hours for that game were sampled with a higher probability, while those with more extensive prior submissions were sampled less frequently.

In this approach, we ensured that the recorded videos were of high quality, expert gameplay.

\subsection{Text Annotation}\label{Appendix:text-annotation}
Here we show the prompt we used for commercial VLM to do text annotations
\begin{lstlisting}
The prompt used for video text annotation is 

You are an expert gaming tutor and video analyst.

─────────────────────────
I. MACRO INSTRUCTIONS
─────────────────────────
A. Definition  
    A MACRO INSTRUCTION is an event that is explicitly listed in the "GAME-FOCUS RULE" table (Section II) or selects the player’s next objective or area and passes **all three** checks: 
   1. Changes the long-term goal or destination (≥ 1 new room, zone, target, phase, or tactic).  
   2. Without it the player would wander or be unsure what to pursue next.  
   3. When applicable, cites a **distinct visual landmark** beginners can recognise (door colour, sign, glowing platform, giant tree, scoreboard icon, billboard, unique prey, ball aura, etc.).  

B. Start & finish timestamps  
   • `"start"` = first frame where the objective becomes clear.  
   • `"end"`   = frame where the objective is satisfied; use `null` if unfinished when the video ends.

─────────────────────────
II. GAME-FOCUS RULE
─────────────────────────
When deciding whether an event is *strategic enough* to log as a macro, follow the table below.  
Skip events that don’t match the focus for that game family.

| Game family                    | Focused macro events (examples)                                                                   |
|--------------------------------|---------------------------------------------------------------------------------------------------|
| **Growth-Arena**  (Roblox *Be a Snake*, *Be a Shark*, *Be a Tornado*) |  • Switching strategy to avoid larger snake/shark/tornado or ambush a smaller snake/shark/tornado ("**Eat the small snake**; **avoid the big shark**"). Use the numbers on top of each snake/shark/tornado's head to indicate the size. • Activating or chasing a power-up landmark (“Dash to the **gold lightning-bolt boost**”). |
| **FPS**  (*Doom*, *Doom2*, *Quake*, *Call of Duty Series*, *Left 4 dead*, etc) | • Reaching or unlocking key areas (“Collect the **red skull key** from the altar”, "press switch on the wall to open the door"). • Navigating path choices (“Enter the **bronze-framed doorway with a skull carving**, **move to the entry with red mark**"”). • Securing major gear required to finish the level (“Take the **rocket launcher on the blood altar**”).  • Use switch to open the door (“Press the switch on the wall to open the door”). • Switch weapons or obtain weapons ("Use axe", "Use gun"). • Pay attention to what the weapon the player is holding, if it was switched then note it down ("switch to gun")|
| **Arena Ball / Dodgeball**  (Roblox *Blade Ball*, *Death Ball*, etc) | • Choosing a safe position or platform (“Jump onto the **neon cube just outside the danger zone**”). • Engaging the ball intentionally (“Charge the **glowing ball with a spin-slash**”). • Triggering or acquiring a special ability landmark (“Grab the **purple deflect power-up circle**”). |
| **Racing**  (*Need for Speed*, etc.) | • Major route decision or shortcut (“Drift through the **billboard shortcut into the alley**”). • Scheduled pit stop or repair landmark (“Pit in at the **blue-neon service lane**”). • Switching race tactics (“Switch to a **draft-and-slingshot strategy behind the lead car**”). |
| **Survival**  (*Natural Disaster Survival*, etc.) | • Major route decision to escape or survive (“Run to the **elevator shaft**”).|

If a game isn’t listed, map it to the closest family.

─────────────────────────
III. GAME-IGNORE RULE
─────────────────────────
When deciding whether an event is **NOT** *strategic enough* to log as a macro, follow the table below.  

| Game family                    | SKIP macro events (examples)                                                                   |
|--------------------------------|---------------------------------------------------------------------------------------------------|
| **Growth-Arena**  (Roblox *Be a Snake*, *Be a Shark*, *Be a Tornado*) |  • Eating small prey from the map (e.g., "Eat the small fog") | 
| **Survival**  (*Natural Disaster Survival*, etc.) | • IGNORE the DISASTER WARNING caption.|

If a game isn’t listed, map it to the closest family.

─────────────────────────
IV. EVENT-FREQUENCY RULES
─────────────────────────
A. Global cool-down  ≥ 8 s between macros (unless a genuine strategy shift appears sooner).  
B. Continuous-action filter (Growth-Arena, racers)  
   • Record a macro only when the agent changes *overall tactic* or *target landmark* (giant pellet, boss prey, pit stop…).  
   • Ignore repetitive minor events.

─────────────────────────
V. DIVERSITY RULE
─────────────────────────
After accuracy is guaranteed, vary verbs, landmark phrases, sentence forms. 
Try to avoid using the same instruction, use diverse description to describe the same event.
Never invent events just to add variety—accuracy outranks diversity.

─────────────────────────
VI. AVOID RULE
─────────────────────────
Avoid using the same instruction to describe the same event, try to be as diverse in wording as possible.
Avoid using directions in the instruction, such as turn right to xxx, turn left to xxx, etc.

─────────────────────────
VII. ON-SCREEN TEXT RULE
─────────────────────────
Capture any explicit text hint/objective that appears on screen.  
Quote it exactly once and log it as a macro if it meets the four checks.

─────────────────────────
VIII. WORKFLOW
─────────────────────────
1. Watch the video once; identify the game and its win condition(s).  
2. **Narrative** – third-person past tense, focus on strategy and playstyle.
3. **Macro list** – array of objects:  
{
"start": "MM:SS",
"end": "MM:SS | null",
"instruction": "<imperative sentence>"
}

─────────────────────────
IX. OUTPUT  (JSON only – no markdown)
─────────────────────────
{
"narrative": "<string>",
"macro_instructions": [
 {
   "start": "00:12",
   "end":   "00:25",
   "instruction": "Enter the bronze-framed doorway with a skull carving"
 },
 {
   "start": "00:34",
   "end":   "00:57",
   "instruction": "Take the corridor lit by flickering red lights"
 },
 {
   "start": "00:01",
   "end":   "00:02",
   "instruction": "Switch to axe"
 },
 {
   "start": "00:52",
   "end":   "00:54",
   "instruction": "Run through without engaging with enemy"
 },
 {
   "start": "01:01",
   "end":   "01:02",
   "instruction": "Change the weapon from gun to fist"
 },
]
}
\end{lstlisting}

\subsection{Simple Environment}\label{Appendix:simple-env}

We constructed some simple games for automated testing with Godot. 

\begin{figure}[H]
    \centering
    \begin{subfigure}{0.465\textwidth}
        \includegraphics[width=\linewidth]{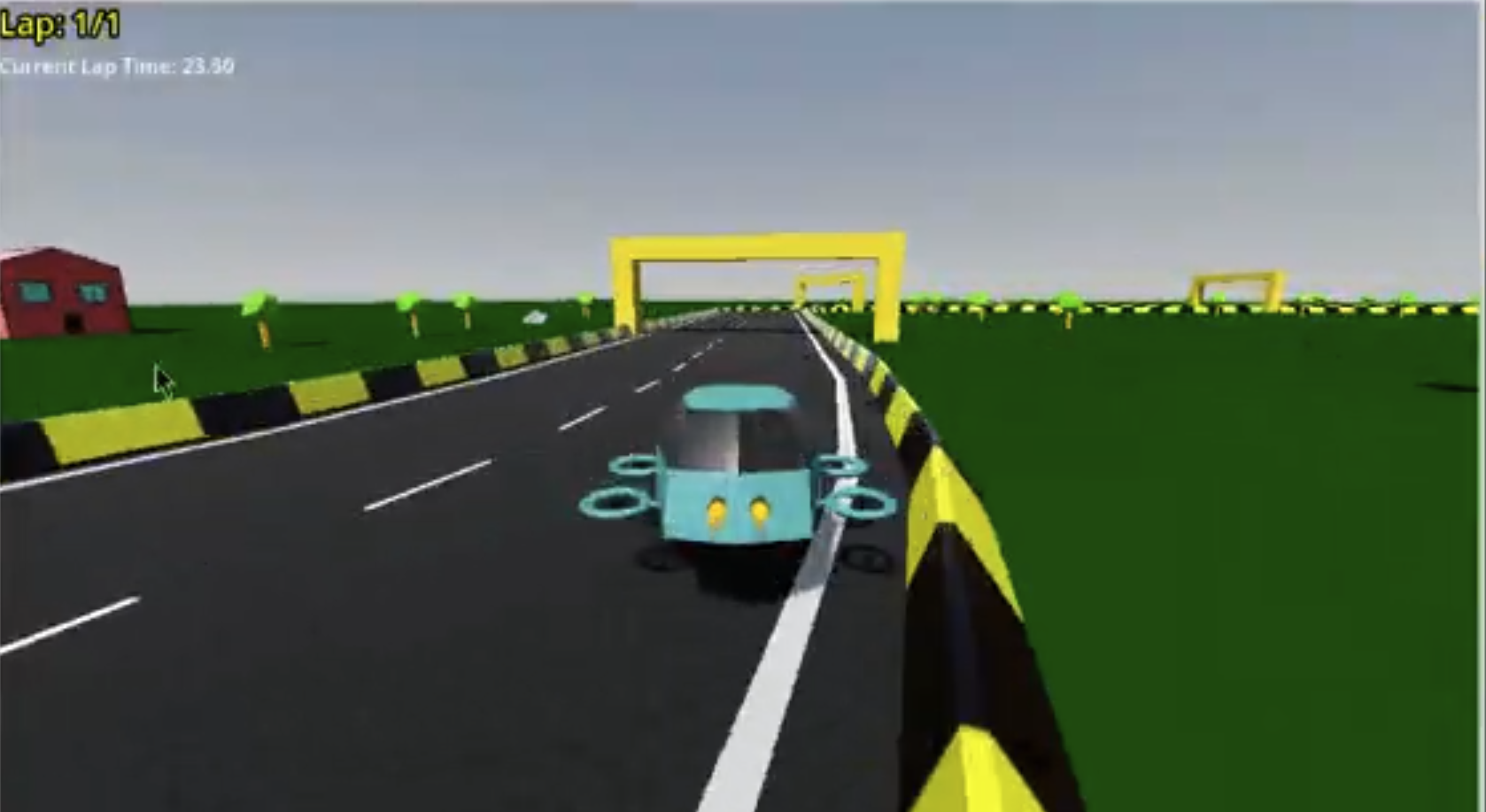}
        \subcaption[]{Hovercraft environment: the car is looping in a fixed racing road. We evaluate the model by measuring how much time it takes to finish one loop}
    \end{subfigure}
    \begin{subfigure}{0.45\textwidth}
        \includegraphics[width=\linewidth]{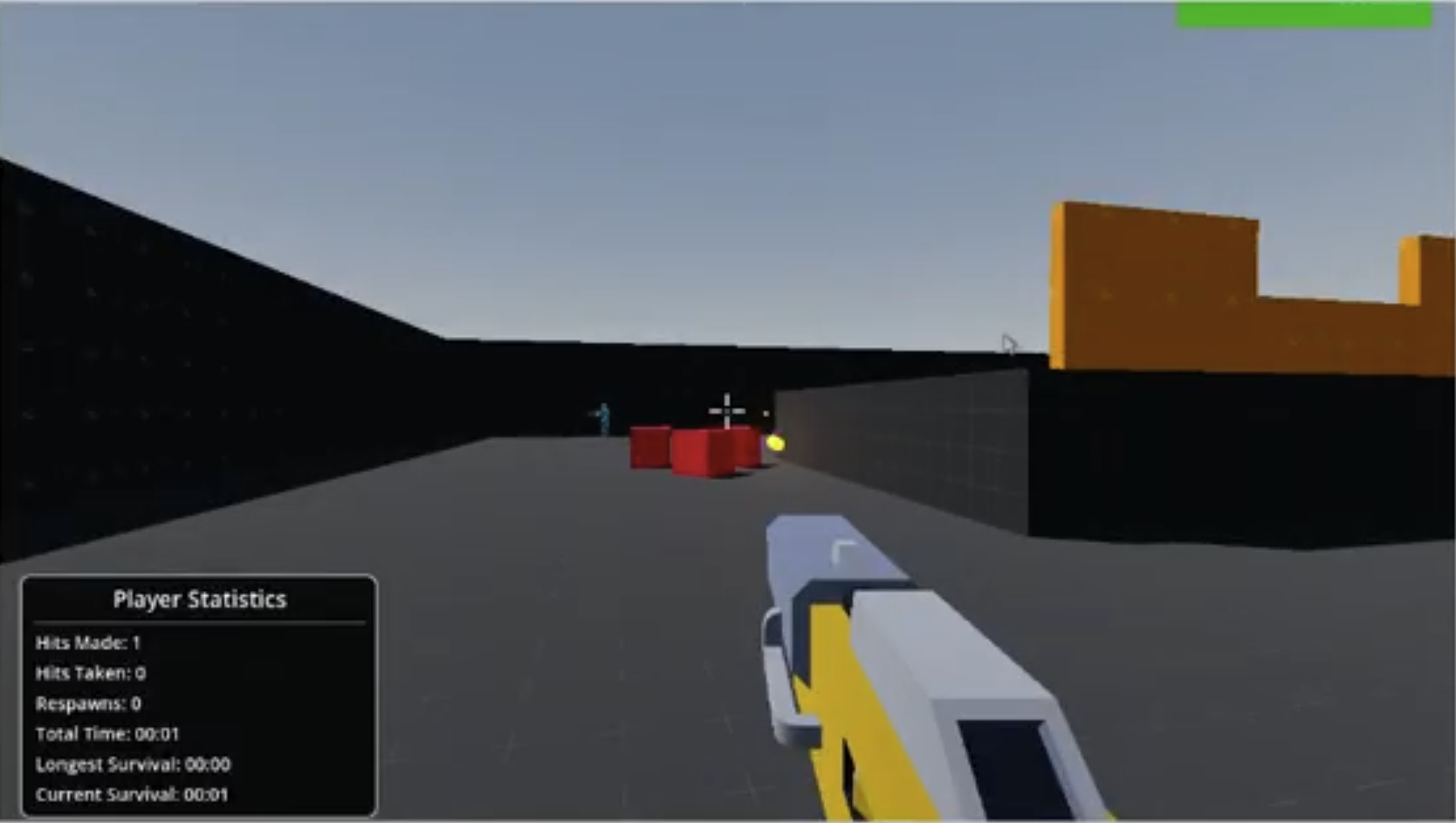}
        \subcaption[]{Simple-FPS environment: a simple fps in a static map. We evaluate the model by counting the number of hit enemies minus the hits taken. }
    \end{subfigure}
\end{figure}

\section{Unlabeled dataset}\label{sec:unlabeled}

\subsection{Unlabeled data distribution}
In addition to annotated gameplay data, we curate a large corpus of unlabeled gameplay videos from public sources \citep{fan2022minedojo, baker2022video}.
While such data are abundant, they present several challenges: highly variable quality, interleaving with non-game content, different resolutions and frame rates, and the absence of corresponding action labels.

We curate unlabeled gameplay trajectories using a two-stage filtering pipeline based on commercial VLMs. In the first stage, videos are filtered using metadata signals such as titles, descriptions, topics, and thumbnails (when available). A VLM is queried to assess relevance to a predefined set of game-related queries. In the second stage, the full video content is processed to segment and remove non-gameplay intervals. To balance cost and filtering accuracy, this step is performed on temporally downsampled video.

\begin{figure}[H]
    \centering
    \includegraphics[width=1.0\linewidth]{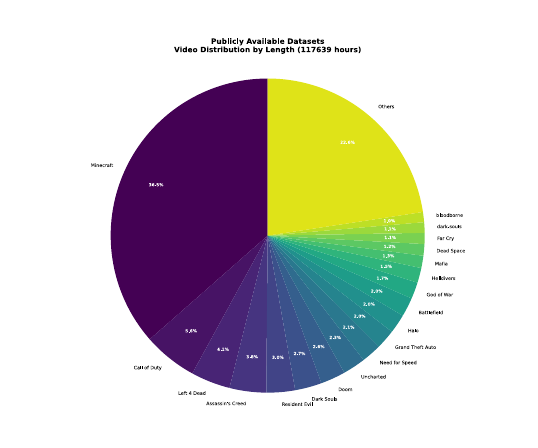}
    \caption{Distribution of games in the unlabeled dataset.}
    \label{fig:unlabeled-data}
\end{figure}

We run queries covering a broad range of popular game titles to collect diverse gameplay footage. The distribution of video hours in the resulting unlabeled dataset is shown in Figure~\ref{fig:unlabeled-data}.

Note that we did not release this dataset as we do not hold the copyright.

\section{Policy Model}

\subsection{Image Tokenizer}
\label{Appendix:frozen}

\begin{figure}[H]
    \centering
    \begin{subfigure}{0.45\textwidth}
        \includegraphics[width=\linewidth]{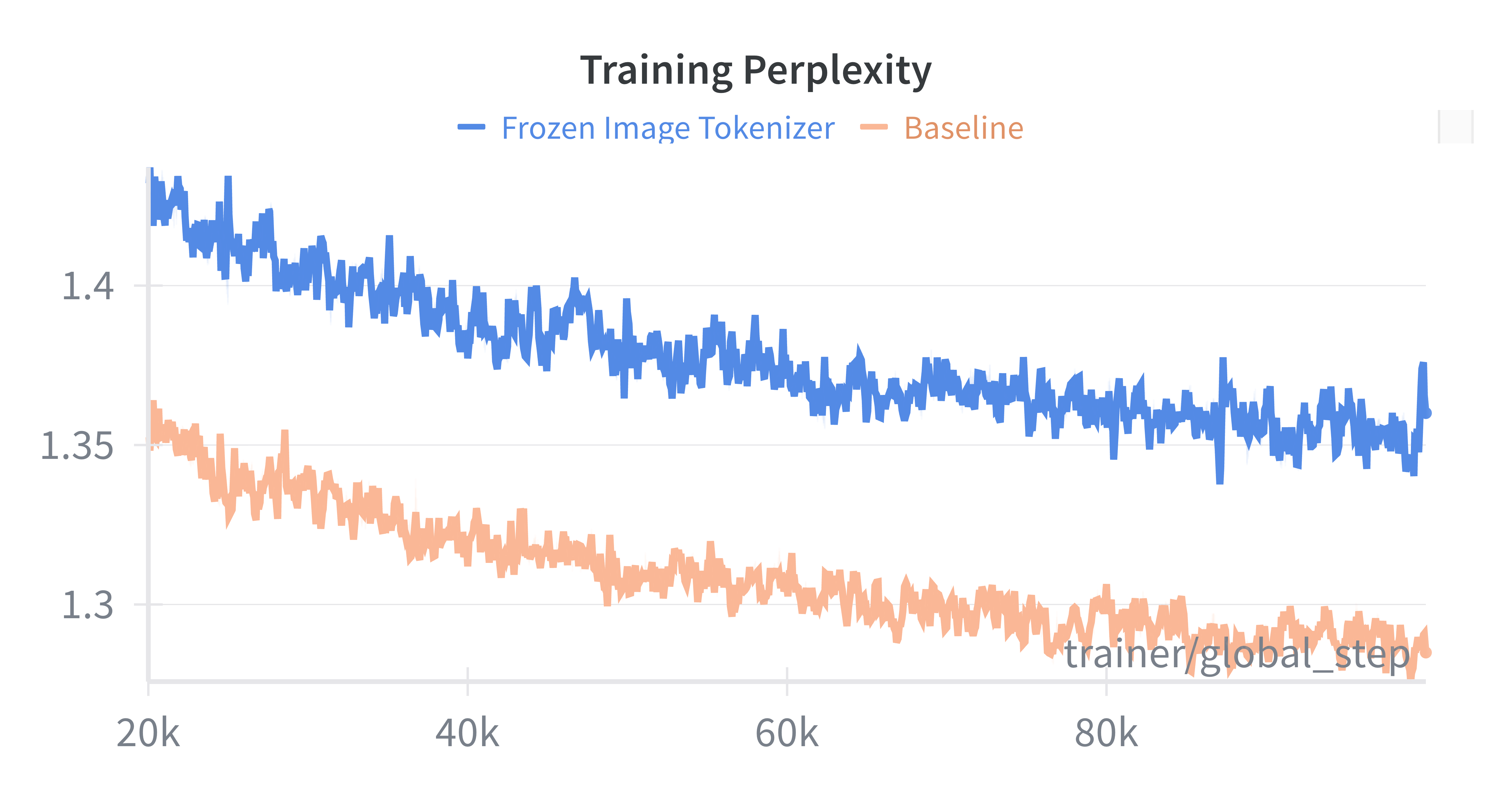}
        \subcaption{Training perplexity with frozen vs.\ unfrozen tokenizer}
    \end{subfigure}
    \hfill
    \begin{subfigure}{0.45\textwidth}
        \includegraphics[width=\linewidth]{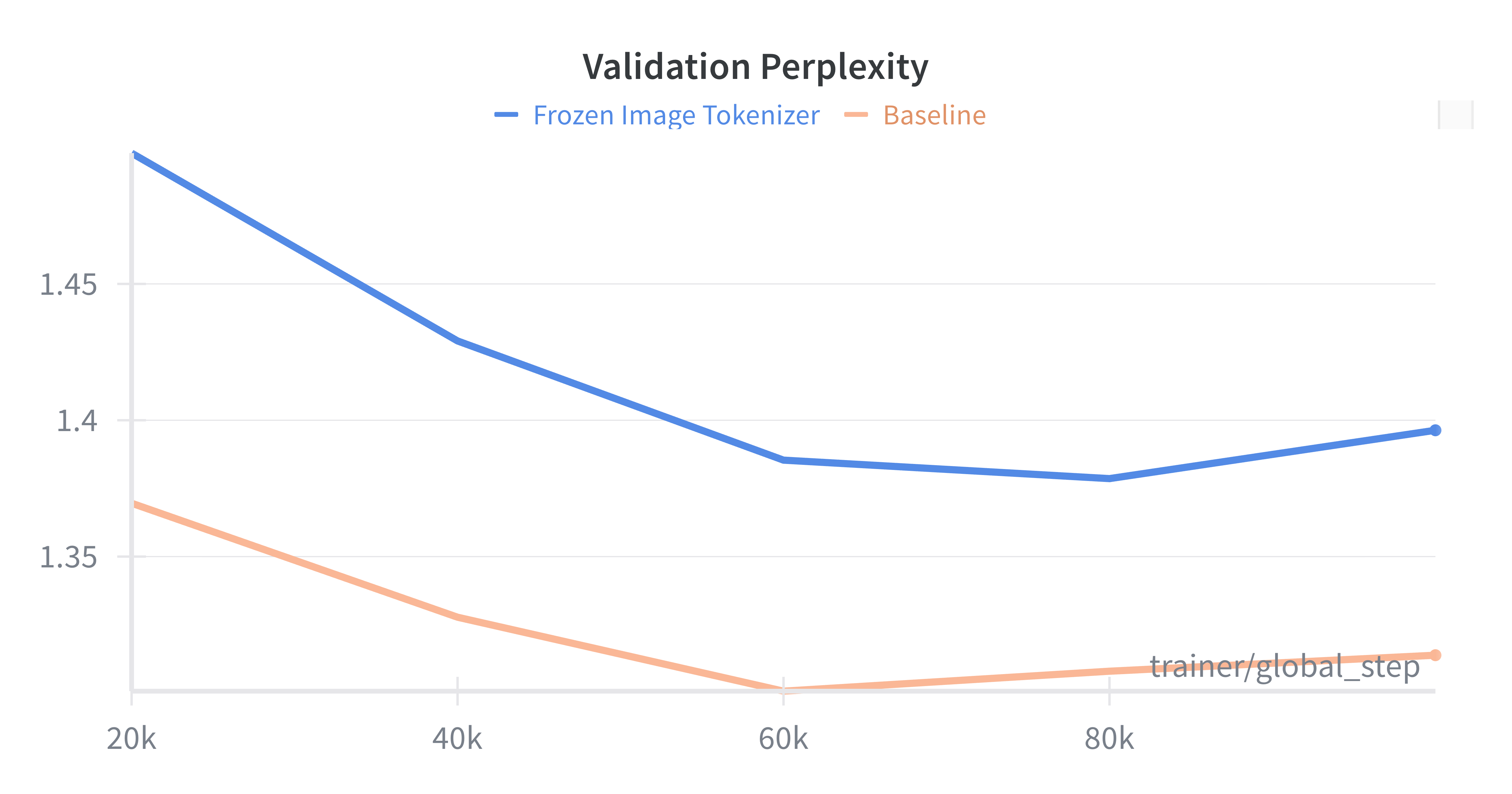}
        \subcaption{Validation perplexity with frozen vs.\ unfrozen tokenizer}
    \end{subfigure}
    \caption{Effect of training the image tokenizer jointly with the policy model. As might expected, training the image tokenizer during model training significantly reduce the training and validation perplexity.}
    \label{fig:frozen-tokenizer}
\end{figure}

Figure~\ref{fig:frozen-tokenizer} compares training and validation perplexity when the image tokenizer is either frozen or trained jointly with the policy model. Jointly training the image tokenizer consistently yields lower perplexity on both training and validation sets, indicating that adapting the visual representation is critical for achieving strong downstream policy performance.

\section{Evaluation}

\subsection{Training details} \label{appendix:training}

All models were trained using automatic mixed-precision training \citep{micikevicius2017mixed}, with model parameters stored in \texttt{float32} and activations in \texttt{bfloat16}, except for RMSNorm layers, which were computed in \texttt{float32}. Common techniques such as z-loss and $K, Q$ norm \citep{rybakov2024methods} were applied to stabilize the training. Training was performed on 8$\times$NVIDIA H100 GPUs. All inference experiments are conducted on a single NVIDIA RTX 5090 GPU; an additional NVIDIA RTX 5080 GPU was used for game rendering to ensure inference ran at a consistent speed. Unless otherwise specified, all experiments share the same set of hyperparameters, which are reported in Appendix~\ref{Appendix:hyperparameter}. We apply data augmentation during training, which we find to be indispensable for mitigating the training--inference gap.

\subsection{Camera setup}\label{Appendix:camera-setup}
We list the in-game camera setup below:
\begin{itemize}
    \item \textbf{Roblox:} Camera sensitivity set to 0.52.
    \item \textbf{Quake:} Mouse sensitivity set to 3.5; smoothing disabled.
    \item \textbf{DOOM:} Smoothing set to 2$\times$; look sensitivity 22\%, move sensitivity 22\%.
\end{itemize}

\subsection{Hyperparameters}
\label{Appendix:hyperparameter}

Table~\ref{tab:hyperparameters} summarizes the hyperparameters used for training the policy model. Batch sizes are chosen to maximize GPU utilization for each model scale. We perform a limited sweep over learning rates \(\{1\times10^{-4}, 3\times10^{-4}, 3\times10^{-5}\}\) on the full dataset and select the best-performing value for each model size, which is then fixed for all remaining experiments. We also list the hyperparameter for data augmentations at Table~\ref{tab:data-augmentation}.

\begin{table}[!t]
\centering
\small
\begin{tabular}{ll}
\toprule
\textbf{Parameter} & \textbf{Value} \\
\midrule
Batch size & $5$ (1.2B), $8$ (600M), $8$ (300M), $10$ (150M) \\
History length & 200 \\
Frame resolution & $192 \times 192$ \\
Text tokenizer & EmbeddingGemma \citep{vera2025embeddinggemma} \\
Image tokenizer & EfficientNet-B0  \\
Number of image tokens & 1 \\
\midrule
\multicolumn{2}{l}{\textbf{Transformer Backbone}} \\
\midrule
Number of layers & 
$10$ (150M), $20$ (300M), $9$ (600M), $22$ (1.2B) \\
Hidden dimension & 
1024 (150M, 300M), 2048 (600M, 1.2B) \\
Query heads & 16 \\
Key--value heads & 16 \\
\midrule
\multicolumn{2}{l}{\textbf{Action Decoder}} \\
\midrule
Number of layers & 3 \\
Query heads & 8 \\
Key--value heads & 8 \\
Hidden dimension & Same as transformer hidden dimension  \\
\midrule
\multicolumn{2}{l}{\textbf{Optimizer (AdamW)}} \\
\midrule
Learning rate & $1\times10^{-4}$ (150M, 300M, 600M), $3\times10^{-5}$ (1.2B) \\
Weight decay & $1\times10^{-4}$ \\
$\beta_1$ & 0.9 \\
$\beta_2$ & 0.999 \\
\bottomrule
\end{tabular}
\caption{Architecture and training hyperparameters across model scales.}
\label{tab:hyperparameters}
\end{table}

\begin{table}[t]
\centering
\caption{Data augmentation parameters.}
\begin{tabular}{l c p{8cm}}
\hline
Augmentation & $p$ & Parameters \\
\hline
Spatial transform & 1.0 & Rotation angle $\sim \mathcal{U}(-3^\circ, +3^\circ)$ \\
Color perturbation & 0.25 & Brightness, contrast, saturation, hue shifts $\sim \mathcal{U}(0.0, 0.2)$ \\
Planckian jitter & 0.25 & Illumination preset $\in \{6, 12, 18, 24\}$ \\
ISO noise & 0.1 & Color shift $\sim \mathcal{U}(0.01, 0.2)$; intensity $\sim \mathcal{U}(0.1, 0.6)$ \\
Random blur & 0.2 & Gaussian: $k \in [3,7], \sigma \in [0.1,2.0]$; or Motion: $k \in [3,7]$, angle $\in [0^\circ,360^\circ]$, direction $\in [-1,1]$ (Gaussian w.p. 0.5) \\
Sharpening & 0.15 & Sharpness factor $\sim \mathcal{U}(0.5, 1.5)$ \\
Translation & 0.25 & Horizontal shift up to $3\%$ of image width \\
\hline
\end{tabular}
\label{tab:data-augmentation}
\end{table}

\subsection{Human Evaluation}\label{Appendix:human-eval}
We report the raw human evaluation scores for DOOM and Quake checkpoints in Table~\ref{tab:raw_scores}. For each model, three checkpoints are evaluated, and each checkpoint is run three times, resulting in 18 total runs.

Human evaluators assessed model quality by counting the occurrences of the following issues during gameplay: (1) colliding with walls (Wall); (2) shooting into the air (Shoot Air); (3) missing targets, including items or enemies (Miss); (4) non–human-like behavior, such as repeating loops or moving backward (Non-human Move); (5) remaining idle (Idle); and (6) camera shaking or jitter (Jitter). Counts were normalized by video length and multiplied by 100 to obtain the final score, with lower values indicating better performance.

Tasks are denoted by the capital letter of the game, followed by the checkpoint number and repetition index (e.g., D1-1 denotes the first run of the first checkpoint in DOOM). Note that Be a Shark and Hypershot are not included in this rubric, as their online environments are dynamic and cannot be reliably controlled across runs.

\begin{longtable}{lccccccccc}
\label{tab:raw_scores} \\
\toprule
\textbf{Task} & \textbf{Model} & \textbf{Wall} & \textbf{Shoot Air} & \textbf{Miss} & \textbf{Non-human Move} & \textbf{Idle} & \textbf{Jitter} & \textbf{Duration} & \textbf{Score} \\
\midrule
\endfirsthead
\toprule
\textbf{Task} & \textbf{Model} & \textbf{Wall} & \textbf{Shoot Air} & \textbf{Miss} & \textbf{Non-human Move} & \textbf{Idle} & \textbf{Jitter} & \textbf{Duration} & \textbf{Score} \\
\midrule
\endhead
\bottomrule
\endfoot
\multirow{4}{*}{D1-1} & 1.2B & 1 & - & - & - & - & - & 28 & 3.57 \\*
 & 600M & 1 & - & - & - & - & - & 24 & 4.17 \\*
 & 300M & - & - & - & 4 & 3 & - & 102 & 6.86 \\*
 & 150M & 3 & - & 1 & 1 & - & - & 109 & 4.59 \\
\midrule
\multirow{4}{*}{D1-2} & 1.2B & 1 & 1 & - & - & - & - & 54 & 3.70 \\*
 & 600M & - & 1 & - & 1 & 1 & - & 74 & 4.05 \\*
 & 300M & - & - & 1 & - & - & - & 30 & 3.33 \\*
 & 150M & 2 & - & 2 & - & - & 1 & 91 & 5.49 \\
\midrule
\multirow{4}{*}{D1-3} & 1.2B & 2 & - & - & 1 & - & - & 92 & 3.26 \\*
 & 600M & 3 & - & 1 & - & - & - & 59 & 6.78 \\*
 & 300M & - & - & - & - & 1 & 1 & 62 & 3.23 \\*
 & 150M & - & 1 & 1 & - & - & - & 38 & 5.26 \\
\midrule
\multirow{4}{*}{D2-1} & 1.2B & 1 & - & - & - & - & - & 29 & 3.45 \\*
 & 600M & - & - & - & 1 & 1 & - & 50 & 4.00 \\*
 & 300M & 3 & 1 & 1 & - & 1 & 1 & 83 & 8.43 \\*
 & 150M & - & - & 1 & - & - & - & 28 & 3.57 \\
\midrule
\multirow{4}{*}{D2-2} & 1.2B & 2 & - & - & - & - & - & 28 & 7.14 \\*
 & 600M & - & 1 & - & - & - & - & 28 & 3.57 \\*
 & 300M & 3 & 1 & - & - & - & - & 84 & 4.76 \\*
 & 150M & 1 & - & - & - & - & - & 21 & 4.76 \\
\midrule
\multirow{4}{*}{D2-3} & 1.2B & 6 & - & - & 1 & - & - & 128 & 5.47 \\*
 & 600M & 2 & - & - & - & - & - & 58 & 3.45 \\*
 & 300M & 1 & 1 & - & 1 & - & - & 84 & 3.57 \\*
 & 150M & 2 & - & 1 & 1 & - & - & 32 & 12.50 \\
\midrule
\multirow{4}{*}{D3-1} & 1.2B & - & - & 2 & 2 & 1 & - & 54 & 9.26 \\*
 & 600M & 3 & 1 & - & 3 & - & 1 & 125 & 6.40 \\*
 & 300M & 5 & 1 & - & 1 & 1 & - & 125 & 6.40 \\*
 & 150M & 3 & 1 & 1 & 3 & 1 & - & 118 & 7.63 \\
\midrule
\multirow{4}{*}{D3-2} & 1.2B & 1 & 1 & - & 3 & - & - & 102 & 4.90 \\*
 & 600M & 1 & - & - & 1 & - & 1 & 45 & 6.67 \\*
 & 300M & 1 & - & 1 & 2 & 1 & - & 95 & 5.26 \\*
 & 150M & 3 & - & 1 & - & - & 2 & 96 & 6.25 \\
\midrule
\multirow{4}{*}{D3-3} & 1.2B & 2 & - & - & - & - & - & 54 & 3.70 \\*
 & 600M & 4 & 1 & 1 & 3 & - & - & 140 & 6.43 \\*
 & 300M & 3 & - & - & 2 & 1 & - & 128 & 4.69 \\*
 & 150M & 2 & - & 1 & - & 1 & - & 66 & 6.06 \\
\midrule
\multirow{4}{*}{Q1-1} & 1.2B & 1 & - & 1 & - & - & - & 41 & 4.88 \\*
 & 600M & - & 2 & - & - & - & 1 & 58 & 5.17 \\*
 & 300M & 4 & - & - & 3 & - & - & 60 & 11.67 \\*
 & 150M & - & - & - & 1 & - & - & 27 & 3.70 \\
\midrule
\multirow{4}{*}{Q1-2} & 1.2B & - & 1 & - & - & - & - & 34 & 2.94 \\*
 & 600M & 2 & - & - & - & - & - & 33 & 6.06 \\*
 & 300M & - & 1 & - & - & - & - & 25 & 4.00 \\*
 & 150M & 1 & 2 & - & 3 & - & - & 85 & 7.06 \\
\midrule
\multirow{4}{*}{Q1-3} & 1.2B & 1 & 1 & - & - & - & - & 43 & 4.65 \\*
 & 600M & 1 & - & - & - & - & - & 26 & 3.85 \\*
 & 300M & 2 & 1 & - & 4 & - & - & 83 & 8.43 \\*
 & 150M & 3 & 1 & - & 1 & - & - & 56 & 8.93 \\
\midrule
\multirow{4}{*}{Q2-1} & 1.2B & - & 1 & 1 & 1 & - & - & 57 & 5.26 \\*
 & 600M & - & 1 & 1 & 3 & - & - & 49 & 10.20 \\*
 & 300M & 2 & - & 2 & 2 & - & - & 66 & 9.09 \\*
 & 150M & - & - & 1 & - & - & - & 27 & 3.70 \\
\midrule
\multirow{4}{*}{Q2-2} & 1.2B & 1 & 1 & 3 & - & - & - & 59 & 8.47 \\*
 & 600M & - & 1 & 1 & - & - & - & 32 & 6.25 \\*
 & 300M & 1 & 1 & - & 2 & - & - & 42 & 9.52 \\*
 & 150M & - & 3 & - & 2 & - & - & 42 & 11.90 \\
\midrule
\multirow{4}{*}{Q2-3} & 1.2B & 1 & - & - & - & - & - & 51 & 1.96 \\*
 & 600M & - & - & - & - & - & - & 21 & 0.00 \\*
 & 300M & 1 & - & - & 1 & - & - & 33 & 6.06 \\*
 & 150M & - & - & - & 3 & - & - & 32 & 9.38 \\
\midrule
\multirow{4}{*}{Q3-1} & 1.2B & - & - & - & - & - & - & 18 & 0.00 \\*
 & 600M & - & - & - & - & - & - & 13 & 0.00 \\*
 & 300M & - & - & 1 & - & - & - & 16 & 6.25 \\*
 & 150M & 2 & - & - & 1 & - & - & 42 & 7.14 \\
\midrule
\multirow{4}{*}{Q3-2} & 1.2B & - & - & - & - & - & - & 10 & 0.00 \\*
 & 600M & 1 & - & - & - & - & - & 23 & 4.35 \\*
 & 300M & - & - & 1 & 2 & - & - & 30 & 10.00 \\*
 & 150M & - & 1 & - & - & - & - & 22 & 4.55 \\
\midrule
\multirow{4}{*}{Q3-3} & 1.2B & - & - & - & 1 & - & - & 26 & 3.85 \\*
 & 600M & - & 1 & - & - & - & - & 24 & 4.17 \\*
 & 300M & 1 & - & - & - & - & - & 26 & 3.85 \\*
 & 150M & 1 & - & - & - & - & - & 13 & 7.69 \\
\end{longtable}

\subsection{Text Instruction Checkpoints}
\label{Appendix:text-instruction}

\begin{figure}[H]
    \centering
    \includegraphics[width=1.0\linewidth]{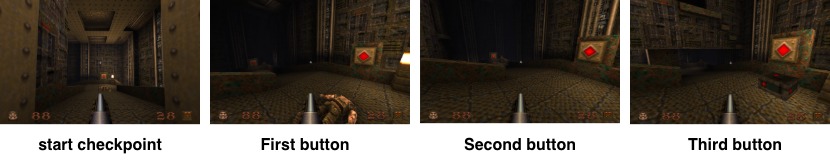}
    \caption{Selected frames from the Quake maze. Three buttons appear along the path, all of which must be pressed to open the door.}
    \label{fig:instruct-following-demo}
\end{figure}

Figure~\ref{fig:instruct-following-demo} shows the checkpoints used for evaluating text-conditioned behavior in Quake. In the maze, there are three red buttons and the player needs to press all of them to open the door. 

\subsection{Scaling Laws}
\label{Appendix:scaling-law}

\begin{figure}[H]
\centering
\includegraphics[width=1.0\textwidth]{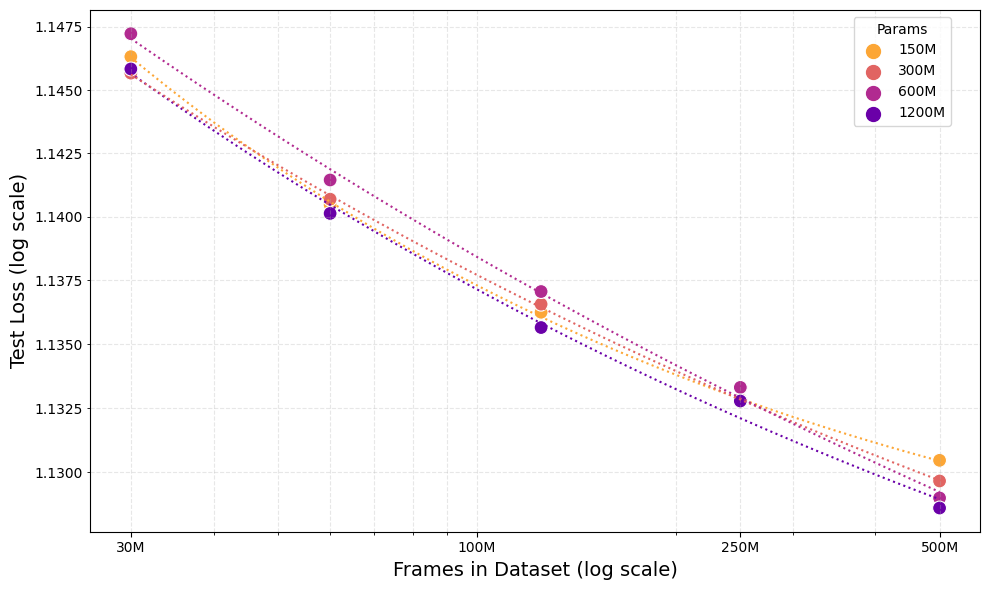}
\caption{Scaling-law curves relating test loss to the number of training frames. All four models are fitted a power-law curve between the data size and test loss, the data exhibits a strong fit to the power-law curve. }
\label{fig:scaling-law-all}
\end{figure}

Figure~\ref{fig:scaling-law-all} presents scaling-law fits for all four model sizes (150M, 300M, 600M, and 1.2B). Across all configurations, test loss closely follows a power-law relationship with respect to the number of training frames. Larger models consistently achieve lower test loss across dataset sizes when dataset size is relatively large.

\begin{figure}[H]
    \centering
    \begin{subfigure}[b]{0.48\textwidth}
        \includegraphics[width=\linewidth]{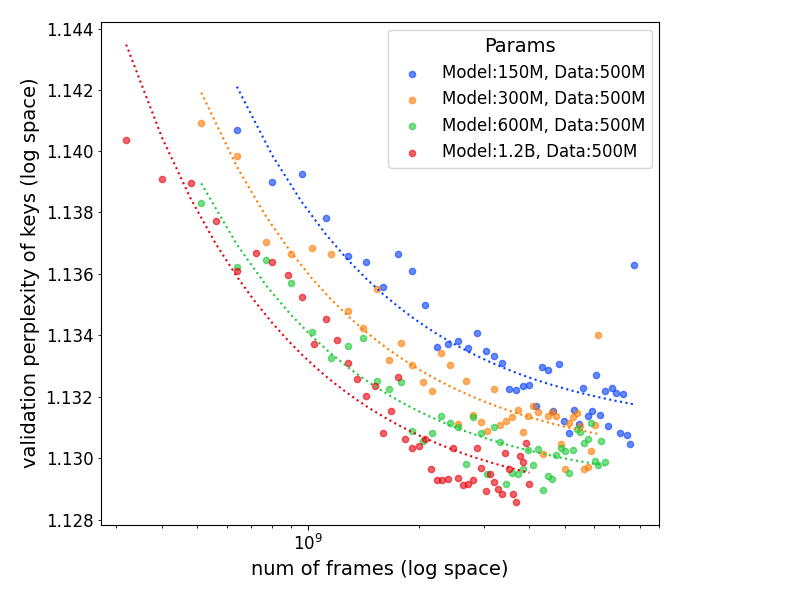}
        \caption{100\% of the dataset}
    \end{subfigure}
    \hfill
    \begin{subfigure}[b]{0.48\textwidth}
        \includegraphics[width=\linewidth]{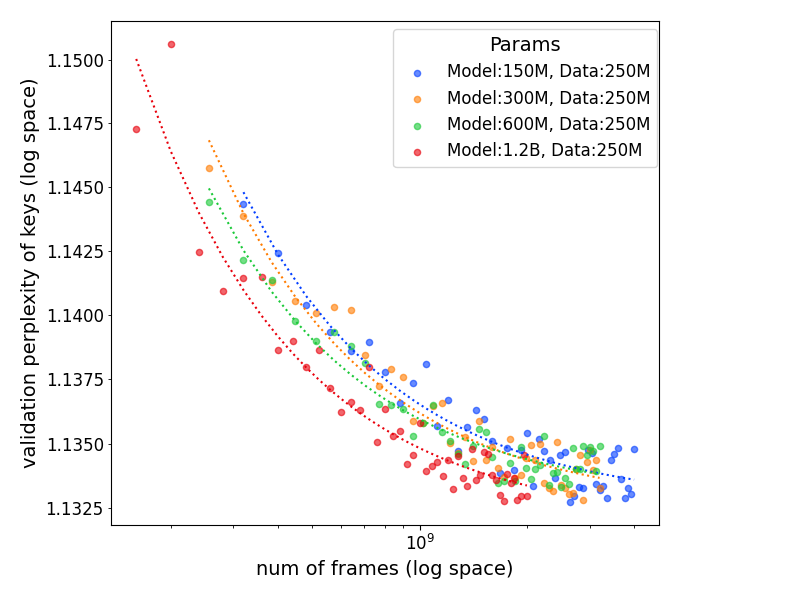}
        \caption{50\% of the dataset}
    \end{subfigure}
    
    \vspace{0.5cm}
    
    \begin{subfigure}[b]{0.48\textwidth}
        \includegraphics[width=\linewidth]{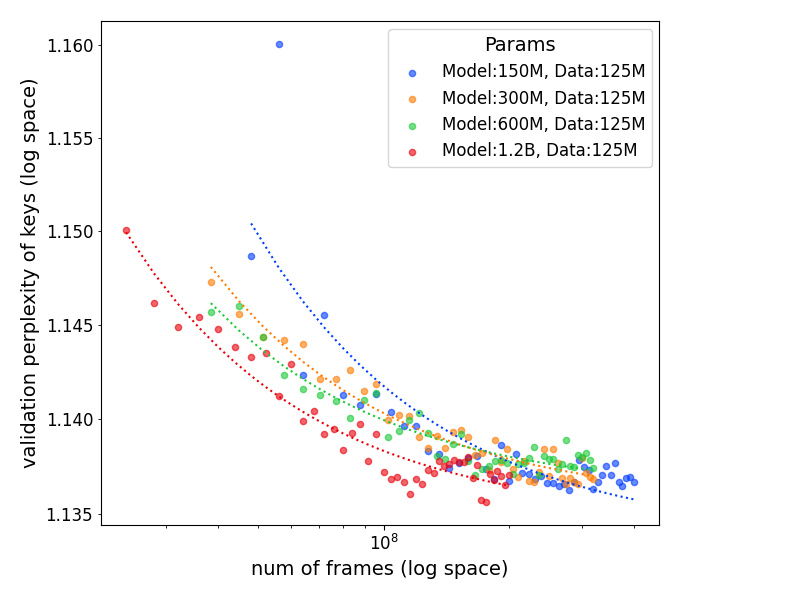}
        \caption{25\% of the dataset}
    \end{subfigure}
    \hfill
    \begin{subfigure}[b]{0.48\textwidth}
        \includegraphics[width=\linewidth]{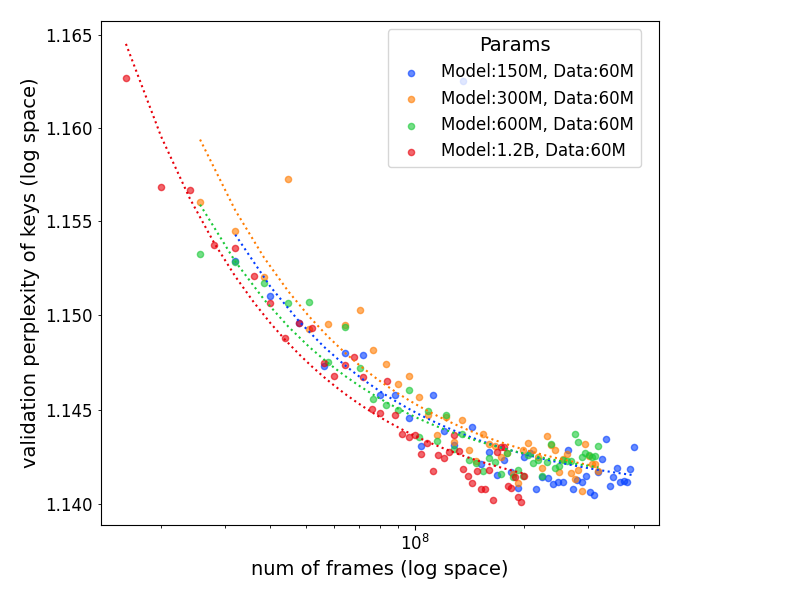}
        \caption{12\% of the dataset}
    \end{subfigure}
    
    \vspace{0.5cm}

    \begin{subfigure}[b]{0.48\textwidth}
        \includegraphics[width=\linewidth]{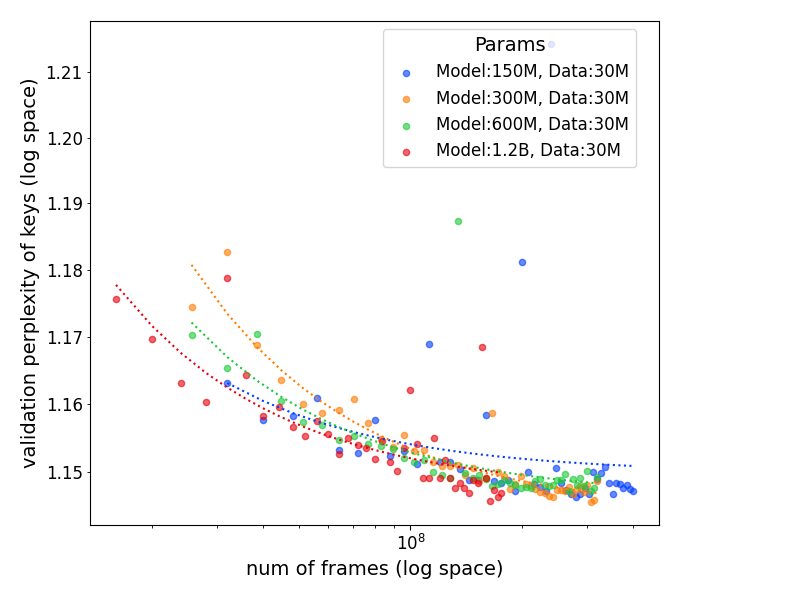}
        \caption{6\% of the dataset}
    \end{subfigure}
    \caption{Test loss as a function of training frames across dataset sizes and model scales. The larger model can leverage the data better when the dataset size is larger, and in general the larger models achieve lower test loss when trained with similar number of frames.}
    \label{fig:scaling-law-appendix}
\end{figure}

Figure~\ref{fig:scaling-law-appendix} further breaks down test loss as a function of training frames across different dataset fractions and model sizes. 

\subsection{Causality Analysis}
\label{Appendix:causality}

\begin{figure}[H]
    \centering
    \includegraphics[width=1.0\linewidth]{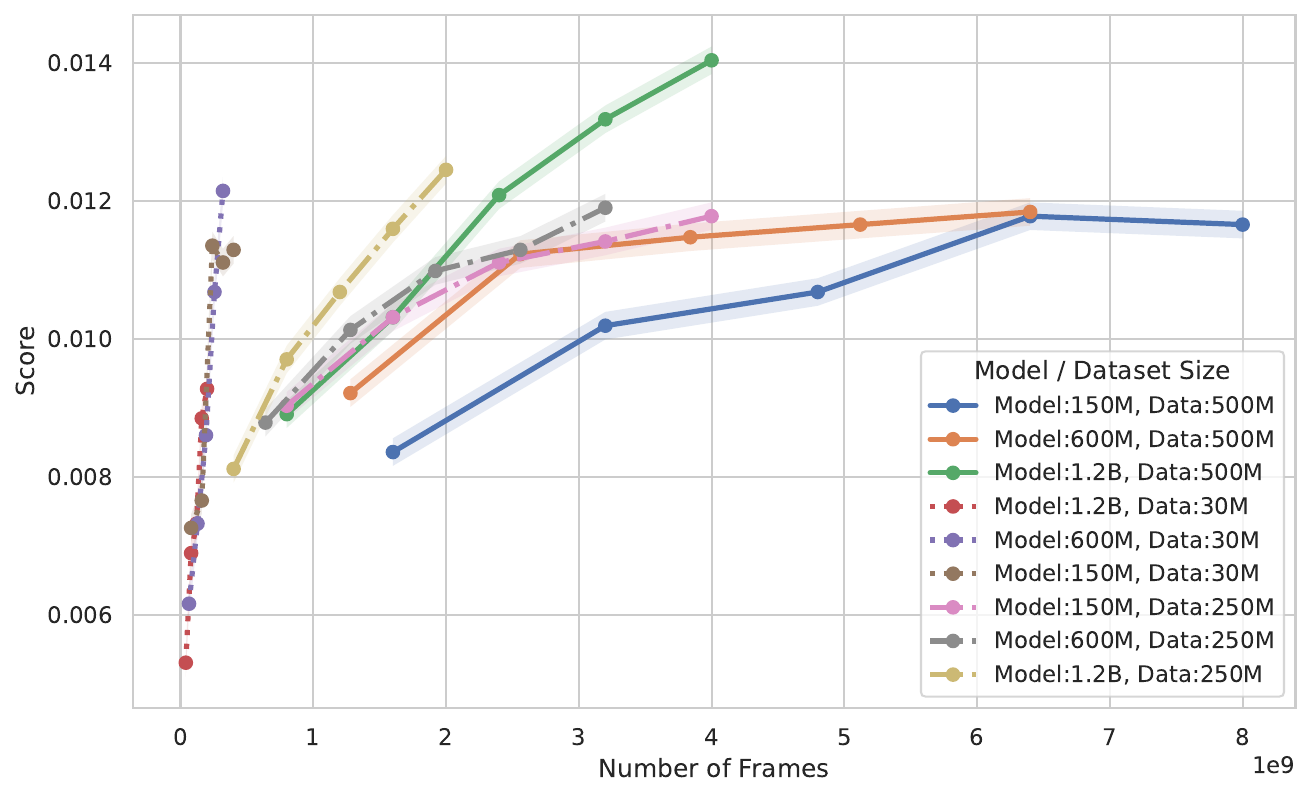}
    \caption{Causality scores as a function of training steps for different model and dataset sizes. Because causality scores generally increase during training, a 600M model trained on 30M samples can exhibit higher causality than the same model trained on 500M samples at intermediate checkpoints. When selecting checkpoints based on lowest test loss, however, the resulting causality trends are consistent with those reported in Section~\ref{sec:causality}.}
    \label{fig:causality-score-appedix}
\end{figure}

Figure~\ref{fig:causality-score-appedix} shows causality scores as a function of training steps for different model and dataset sizes. Because causality scores consistently increase over the course of training, the 600M model trained on the 30M dataset can exhibit a higher causality score than the same model trained on the 500M dataset at certain training steps. However, when we instead select model checkpoints based on the lowest test loss and then evaluate causality, the resulting trends align with those discussed in Section~\ref{sec:causality}.

\subsection{Pretraining with Unlabeled Data} \label{Appendix:pretrain}
Unlabeled gameplay videos are orders of magnitude more abundant than human annotated demonstrations. We therefore leverage an inverse dynamics model (IDM) to convert unlabeled videos into additional training data.

\subsubsection{Inverse Dynamics Model}
Two classes of IDMs have been explored in prior work: \emph{real-action} models that directly predict keyboard and mouse actions \citep{baker2022video}, and \emph{latent-action} models that infer abstract action codes subsequently mapped to real actions \citep{schmidt2023learning, ye2024latent, nvidia2025gr00t}. For simplicity, we adopt the real-action formulation.

Formally, the IDM predicts the action at time $t$ conditioned on the surrounding image sequence:
\[
\tilde{a}_t \sim p_{\mathrm{IDM}}\bigl(a_t \mid o_1, o_2, \ldots, o_T \bigr).
\]
The IDM shares the same architecture as the policy model (Figure~\ref{subfig:model}), with two key differences. First, it does not condition on text or ground-truth action tokens, since environment dynamics are independent of text inputs and the IDM must predict all action labels in a single forward pass rather than autoregressively. Second, the IDM is non-causal and is allowed to attend to future frames, which improves action imputation accuracy. This architectural alignment ensures that improvements to the policy model transfer directly to the IDM.

The IDM was trained using cross-entropy loss on the labeled dataset. To ensure robustness to the diversity of unlabeled data, we applied extensive data augmentation during training. Once trained, the IDM was used to impute actions for the unlabeled gameplay dataset.

\subsubsection{Evaluation of pretrained model}

We trained a 600M model that leverages unlabeled data to study the benefits of large-scale pretraining. We refer to this model as \emph{pretrained-600M}, which is obtained through a three-stage procedure:
(1) training an inverse dynamics model (IDM) on the full labeled dataset;
(2) using the trained IDM to generate pseudo-labels for an unlabeled dataset that is approximately $10\times$ larger than the annotated dataset, followed by pretraining the 600M policy model on this pseudo-labeled data for one epoch; and
(3) fine-tuning the model on the full labeled dataset.

Evaluating the quality of the IDM presents a nontrivial challenge. Since ground-truth action labels are only available for the annotated dataset, quantitative evaluation of the IDM can only be performed on this data, on which the IDM is also trained. As a result, standard evaluation metrics tend to overestimate performance and do not fully reflect the IDM’s effectiveness in its intended deployment setting. In practice, the IDM is primarily used to generate pseudo-labels for unlabeled gameplay videos, whose distribution differs substantially from that of the labeled data.

To account for this distributional gap, we complement quantitative evaluation with manual inspection. Specifically, we sample unlabeled videos and assess the plausibility and temporal consistency of the generated pseudo-labels using human judgment. This qualitative evaluation provides an additional sanity check on whether the IDM produces reasonable action annotations when applied to out-of-distribution data.

\subsubsection{Simple Programmatic Environment}
Similar to the evaluation of other models, we show the evaluation on Godot environment of the 600M pretrained model at Table~\ref{tab:godot-env-pt}. There is incremental changes in the scores when compared to its 600M counterpart. 
\begin{table}[H]
\centering
\small
\rowcolors{2}{gray!10}{white}
\begin{tabular}{lccc}
\toprule
\textbf{Model Size} &
\textbf{Hovercraft $\downarrow$} &
\textbf{Simple-FPS $\uparrow$} &
\textbf{FPS $\uparrow$} \\
\midrule
Pretrained 600M & 38  & 24 & 61 \\
\bottomrule
\end{tabular}
\caption{Performance on Godot-based programmatic environments across model sizes for pretrained 600M model. There is only incremental change when comparing with the 600M scores from Table~\ref{tab:godot-env}.}
\label{tab:godot-env-pt}
\end{table}

\subsubsection{Quantitative Eval}
\begin{figure}[H]
    \centering
    \includegraphics[width=0.8\linewidth]{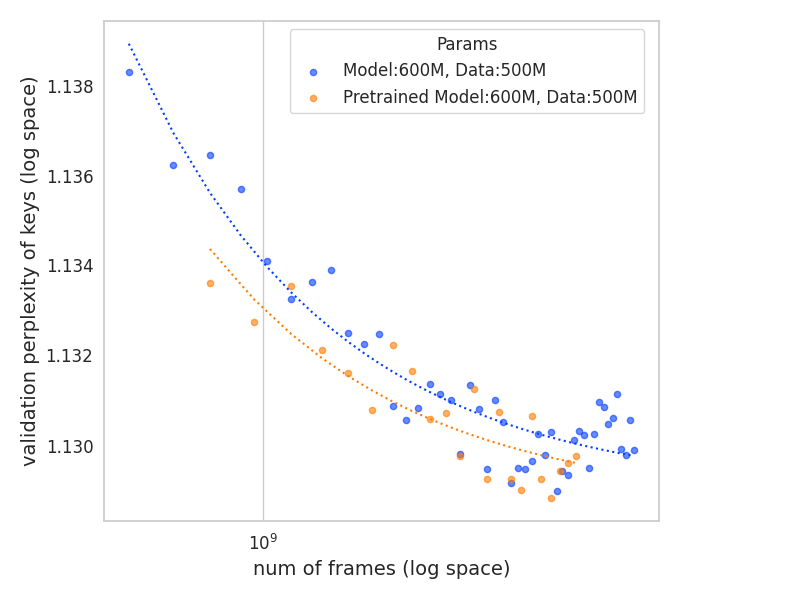}
    \caption{Comparison of test loss between pretrained and label-only 600M models. The pretrained 600M model achieves substantially lower test loss than a model trained solely on labeled data when trained on the same number of frames.}
    \label{fig:pretrain-loss}
\end{figure}

Here, we compare the test loss of the pretrained 600M model with that of a 600M model trained solely on labeled data. As shown in Figure~\ref{fig:pretrain-loss}, the pretrained 600M model achieves substantially lower test loss than the model trained only on labeled data when both are trained on the same number of frames.

\subsubsection{Human Evaluation in Real Environment}
We compare gameplay videos generated by the pretrained 600M model and the 600M model trained using labeled data only. Although the pretrained model achieves lower test loss, its online performance is not significantly preferred over the non-pretrained model. 

We hypothesize that this gap arises because pretraining incorporates a large amount of unrelated video data, which may introduce atypical or environment-specific movements when deployed in a particular game. Such behaviors can appear less human-like, which is a critical factor in human preference during evaluation.

\end{document}